\documentclass{article} % For LaTeX2e
\usepackage{collas2025_conference,times}
\usepackage{easyReview}

\usepackage[utf8]{inputenc} % allow utf-8 input
\usepackage[T1]{fontenc}    % use 8-bit T1 fonts
\usepackage{hyperref}       % hyperlinks
\usepackage{url}            % simple URL typesetting
\usepackage{booktabs}       % professional-quality tables
\usepackage{amsfonts, amsmath, amsthm}       % blackboard math symbols
\usepackage{nicefrac}       % compact symbols for 1/2, etc.
\usepackage{microtype}      % microtypography
\usepackage{xcolor}         % colors

\usepackage{bbm}
\usepackage{graphicx}
\usepackage{wrapfig}
\usepackage{float}
\usepackage{enumitem}
\usepackage{caption}
\usepackage{subcaption}

\usepackage{algorithm, algorithmic}
\usepackage{wrapfig}
\usepackage{cleveref}

\newtheorem{definition}{Definition}

\usepackage{amsmath,amsfonts,bm}

\def\eqref#1{equation~\ref{#1}}
\def\1{\bm{1}}

\DeclareMathAlphabet{\mathsfit}{\encodingdefault}{\sfdefault}{m}{sl}
\SetMathAlphabet{\mathsfit}{bold}{\encodingdefault}{\sfdefault}{bx}{n}

\usepackage{hyperref}
\hypersetup{
    colorlinks=true,
    linkcolor=red,
    filecolor=magenta,
    urlcolor=blue,
    citecolor=purple,
    pdftitle={Overleaf Example},
    pdfpagemode=FullScreen,
    }

\title{Manifold Metric: A Loss Landscape Approach for Predicting Model Performance}

\author{%
  Pranshu Malviya\textsuperscript{1,2,3} \quad
  Jerry Huang\textsuperscript{1,2,4} \quad
  Aristide Baratin\textsuperscript{5} \quad
  Quentin Fournier\textsuperscript{1,2} \quad
  Sarath Chandar\textsuperscript{1,2,3,6} \\ \\
  \textsuperscript{1}Chandar Research Lab \quad
  \textsuperscript{2}Mila – Qu\'{e}bec AI Institute \quad
  \textsuperscript{3}Polytechnique Montr\'{e}al \\
  \textsuperscript{4}Universit\'{e} de Montr\'{e}al \quad
  \textsuperscript{5}Samsung SAIT AI Lab Montreal \quad
  \textsuperscript{6}Canada CIFAR AI Chair
}

\collasfinalcopy 

\begin{document}

\maketitle

\begin{abstract}
    Determining the optimal  model for a given task often requires training multiple models from scratch,  which becomes impractical as dataset and model sizes grow. A more efficient alternative is to expand smaller pre-trained models, but this approach is underutilized due to a limited understanding of its impact on the training dynamics. Existing methods for quantifying this impact have notable limitations, including computation cost.  To address this, we introduce a new perspective based on the loss landscape, which has been shown to contain a manifold of linearly connected minima. Specifically, we propose a metric that estimates the size of this manifold to study the impact of model expansion.  Our experiments reveal a strong correlation between performance gains and our manifold metric, enabling more informed model comparison and offering a first step toward a geometry-driven approach for reliable model expansion. Notably, our metric outperforms other baselines, even when different types of expansion with equivalent number of parameters are applied to a model.
\end{abstract}

\section{Introduction}

The size of a neural network significantly impacts its performance, with several works suggesting that as training datasets grow, scaling models and training them from scratch leads to improved performance~\citep{kaplan2020scaling}. However,  this approach is computationally expensive. A more efficient alternative is to expand smaller pre-trained models, allowing for knowledge transfer~\citep{chen2016net2net, wang2023learning}. This can be particularly valuable in online continual learning,  where a model must retain past knowledge while adapting to distribution shifts~\citep{yoon2017lifelong,ye2023self}. Despite these advantages, model expansion remains poorly understood and,  if done improperly, it can lead to sub-optimal performance, requiring further training~\citep{abdelfattah2021zero}. Establishing a link between expansion strategies and final performance is thus essential to avoid costly inefficiencies.

Several metrics have been proposed to predict the final performance of a model based on partially trained candidates~\citep{white2023neural}. However, as models scale, performance estimations must meet two key criteria: (i) they should be reliable and independent of the optimization process,  and (ii) they should be cost-effective to compute. Existing methods often fail in at least one of these aspects—informative statistics are computationally expensive or require additional training, while cheaper alternatives tend to be unreliable~\citep{white2021powerful}. As models evolve, the need for robust and efficient metrics becomes increasingly critical, yet developing such metrics remains an open challenge.

A promising approach is to leverage the properties of the loss landscape, which has been shown to encode rich structural information about optimization and generalization~\citep{sam,liu2022loss,barannikov2020topological}. The shape of the loss landscape is determined by two key factors: the model's architecture and the training data distribution~\citep{cooper2018loss}, both of which influence final performance~\citep{li2018visualizing,li2023understanding}. 
Analyzing or estimating these geometric properties can provide valuable insights into training dynamics~\citep{sun2020global,kunin2019loss}. This leads us to ask: 
\begin{quote}
	\textit{Can we understand and quantify the impact of model expansion through the lens  of the loss landscape?}
\end{quote}

Empirical findings suggest that geometric properties may hold the answer. \citet{sun2019optimization} observed that the loss landscapes of high-dimensional models exhibit non-trivial properties related to optimization difficulty. One key property, linear mode connectivity (LMC), suggests that minima obtained from different training runs can often be connected via linear or low-loss paths~\citep{garipov2018loss, frankle2020linear, anokhin2020low}. \citet{mirzadeh2020linear} showed that incorporating mode connectivity as constraints over learned minima in continual learning leads to improved performance. Further studies have shown that these connections between minima emerge more robustly when model parameters are permuted in a function-preserving manner~\citep{entezari2021role, ainsworth2022git}. 
%further show the emergence of this connectivity between multiple minima when permuting the model parameters in a function-preserving way.
Importantly, \citet{pmlr-v139-benton21a} and \citet{simsek2021geometry} demonstrated that such minima often reside within a larger manifold of low-loss solutions, highlighting the significance of connectivity in optimization landscapes.

\begin{figure*}[thb!]
	\centering
	\resizebox{\linewidth}{!}{
		\includegraphics[width=0.32\linewidth]{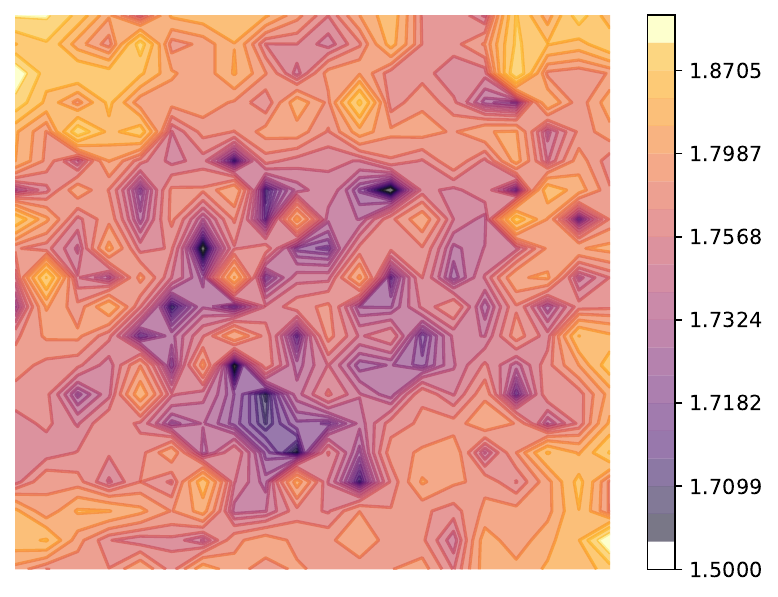}
		\includegraphics[width=0.32\linewidth]{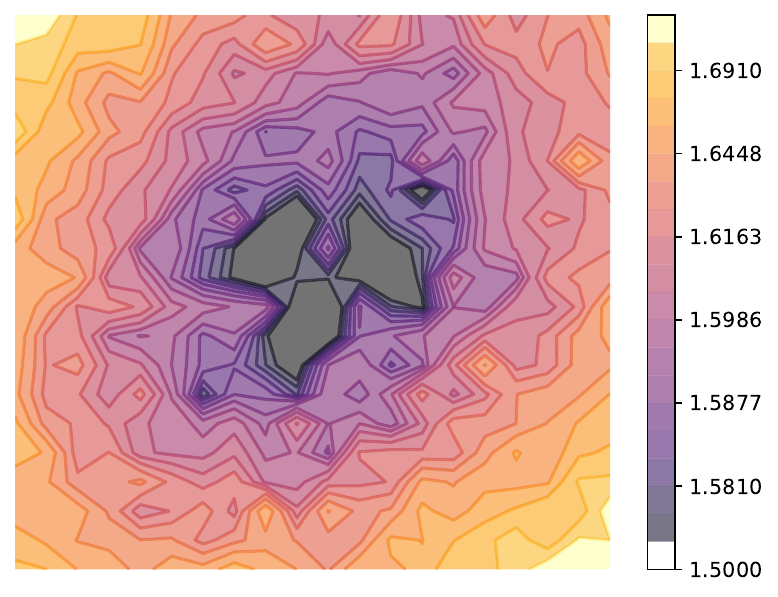}
		\includegraphics[width=0.32\linewidth]{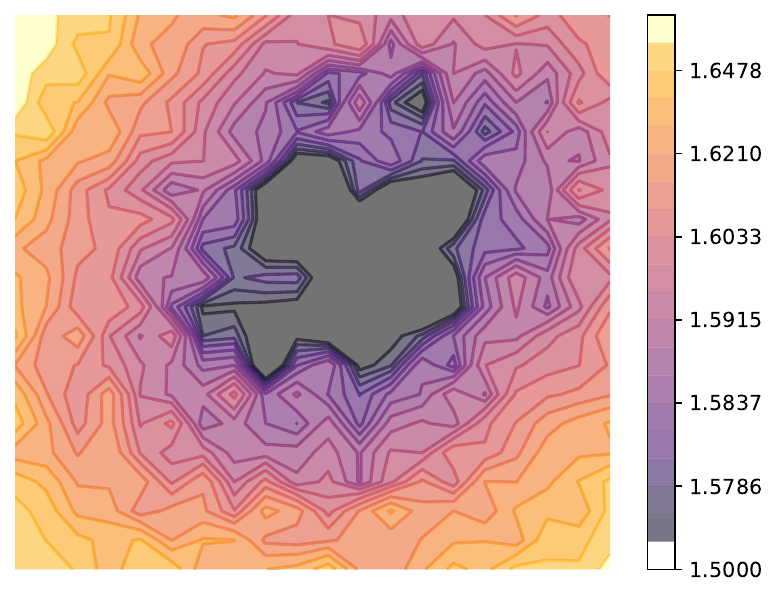}
	}
	\caption{LMC between minima using a MLP with one hidden layer ($20$ units) pre-trained on CIFAR10. The initial model (first) is expanded by increasing the hidden layer size in a function-preserving way and trained further until convergence. Final models after hidden size expansions of $2\times$ (second) and $3\times$ (third) show a reduced loss and enhanced local connectivity between minima.}
	\label{fig:loss_landscape}
\end{figure*}

Building on these insights, we argue that connectivity plays a crucial role in understanding how model expansion influences the loss landscape and the final performance.  As shown in \autoref{fig:loss_landscape},  mode connectivity increases when a small network is expanded along its width. Motivated by this, we propose a metric  that quantifies local connectivity by estimating changes in the size of the minima manifold, providing a novel tool for comparing expansion strategies.

To evaluate our approach, \textcolor{black}{we focus on function-preserving model expansion for a given task and} conduct a series of experiments on various models and benchmarks commonly used to study loss landscapes~\citep{simsek2021geometry, wang2022plateau, zhou2023understanding}. While prior work has focused primarily on multi-layer perceptrons (MLPs) and convolutional neural networks (CNNs), we extend the analysis to Transformers~\citep{Vaswani-2017}, which exhibit distinct optimization characteristics~\citep{yang2021taxonomizing}. Our contributions are as follows:
\begin{itemize}%[leftmargin=12pt,noitemsep,topsep=0pt]
    \item {\bf Loss landscape analysis of model expansion:} We examine how expansion  impacts  the loss landscape and use this information to estimate performance gain in high-dimensional neural networks.
    \item {\bf Empirical validation in image classification:} We demonstrate that our proposed metric is a more robust predictor of performance gains in image classification as it outperforms existing baselines, even when different types of expansion with equivalent parameter counts are applied to a base model. %\textcolor{red}{TODO: mention about the main result -same size }
    \item {\bf Insights from language modeling:} We find that during Transformer pre-training, the mode connectivity significantly reduces as the model is trained after expansion, suggesting that Transformers may not exhibit the same LMC properties as CNNs.
\end{itemize}

\section{Related work}\label{sec:related}

\subsection{Neural Network Expansion}
Expansion can reduce the computational cost of training neural networks by re-using smaller pre-trained models. Many real-world applications are inherently continuous~\citep{douillard2022dytox},  requiring models to increase in capacity to accommodate learning across multiple datasets~\citep{li2019learn}. Accordingly, various expansion strategies have been explored in deep learning. Early approaches by \citet{fahlman1989cascade} and \citet{Gutstein2007KnowledgeTI} added hidden units and layers, respectively, but kept existing parameters frozen, limiting learning to the new parameters. In contrast, \citet{chen2016net2net} introduced Net2net,  a function-preserving expansion method for MLPs and CNNs that that allows existing units to remain trainable. More recently, expansion techniques have been explored for Transformers as well~\citep{gong2019efficient, gu-etal-2021-transformer, wang2023learning, gesmundo2023composable}.
However,, these methods lack a systematic way to assess whether expanding the model will improve performance and provide no guidance on how or by how much to expand. Unlike these prior works, we do not introduce a new expansion method; instead, we introduce a flexible and inexpensive way to easily estimate the impact of expansion on the model's generalization.

\subsection{Finding Optimal Models with Performance Prediction} While finding the optimal model among many is complex, advances have been made. Most notably, neural architecture search (NAS)~\citep{zoph2017neural} automates the process of finding the best-performing model by conducting an automatic search over a space of architecture and ranking them by predicting their performance on a task~\citep{ren2021comprehensive}. Furthermore, numerous techniques have been explored to predict the final performance of a model after expansion. \citet{white2021powerful} introduced a taxonomy that includes the following categories. Model-based methods train several candidate models to construct a dataset to train a parameterized predictor model, thereby requiring large initialization times~\citep{springenberg2016bayesian, siems2020bench, white2021bananas}. Learning curve extrapolation methods rely on partially trained candidate models to predict the final performance and, therefore, exhibit high query times~\citep{ru2020revisiting, zoph2018learning}. Zero-cost proxies provide estimates of the (relative) performance of neural architectures from a small amount of data~\citep{mellor2021neural, lee2018snip, wang2020picking, tanaka2020pruning}. Our method belongs to the zero-cost proxies and uses the local loss landscape for making predictions.

\subsection{Properties of the Loss Landscape} 

Several established metrics characterize the properties of the loss landscape, some of which have been used for NAS~\citep{abdelfattah2021zero, white2021powerful, mellor2021neural, kaur2023maximum, shen2023proxybo}. However, these metrics can be highly dependent on the optimization process and location of parameters. In contrast, metrics that estimate the geometric properties of the loss landscape inherently mitigate this dependency. One such geometric property that has gained interest is manifold size~\citep{kunin2019loss}. In particular, \citet{simsek2021geometry} noted that permutation symmetries create multiple equivalent global minima and that expanding a model in width can connect these minima into a single manifold, potentially facilitating convergence to global minima. 

\citet{frankle2020linear} suggested that higher connectivity between minima is desirable,  as it indicates optimization stability  and has implications for the lottery ticket hypothesis--the idea that neural networks contain sparse subnetworks that can be trained in isolation to achieve better generalization performance~\citep{juneja2022linear, zhang2021lottery}. While the theoretical relationship between generalization and mode connectivity remains an open question~\citep{zhou2023going}, a common observation is that a connected landscape of low-loss solutions corresponds to convex regions, aiding optimization and improving generalization~\citep{neyshabur2020being, mirzadeh2020linear}. We take this connection as motivation for comparing the mode connectivity in candidate models when expanding a pre-trained model~\citep{wu2018sgd, fort2020deep}. However, explicitly addressing this relationship is beyond the scope of this paper.

\section{Methodology}\label{sec:methodology}

This section describes our proposed metric for estimating the benefit of expansion by quantifying changes in the loss landscape, more specifically the manifold size.

\subsection{Linear Mode Connectivity}

Our metric relies on the LMC between minima measured using the loss barrier~\citep{entezari2021role} defined below.

\begin{definition}[Loss Barrier]
	Consider two networks with parameters $\theta_1, \theta_2 \in \mathbb{R}^k$ and a loss function $\mathcal{L}$ evaluated on a dataset $\mathcal{D}$. The \textbf{loss barrier} $b(\theta_1, \theta_2)$ along the linear path between $\theta_1$ and $\theta_2$ is defined as the greatest difference between the loss of the linear interpolation between the two networks and the linear interpolation between their losses:
	\begin{equation}
		\label{eq:barrier}
		b(\theta_1, \theta_2) = \sup_\alpha \mathcal{L}(\alpha\theta_1 + (1 - \alpha)\theta_2) \\ - [\alpha \mathcal{L}(\theta_1) + (1 - \alpha)\mathcal{L}(\theta_2)]
	\end{equation}
	where $\alpha \in [0,1]$ determines the location on the linear path. 
\end{definition}

More intuitively, different configurations exist along the linear interpolation between two network configurations $\theta_1$ and $\theta_2$. If, at any point along this path, the true loss value differs from the hypothetical loss value assuming a linear path, a \textit{barrier} blocks the path between the two network configurations in the loss surface. The networks are said to be linearly connected if their loss barrier is zero. 

\subsection{Manifold Detection}

\citet{entezari2021role} and \citet{ainsworth2022git} proposed search algorithms to find the permutation of hidden units that minimizes the loss barrier (\autoref{eq:barrier}) with the goal of aligning the weights of two independently trained models. In contrast, we propose an algorithm to detect the presence of a manifold within a given model. {We represent the connectivity between minima as a graph where nodes represent distinct models generated by permutations of a base model trained to convergence and whose edges indicate LMC between two permutations with a relaxed linearity constraint.}

For a model parameterized by $\theta \in \mathbb{R}^k$, we obtain a new parameter configuration $\theta_\texttt{perm}$ by randomly permuting neurons in $\theta$ such that both lie in functionally equivalent regions {i.e., $\mathcal{L}(\theta) = \mathcal{L}(\theta_{\texttt{perm}})$}~\citep{kuditipudi2019explaining, pena2023re}. If $T$ represents the set of permutations that produce functionally equivalent parameters to $\theta$, then the permutation function is given by $P: \mathbb{R}^k \times T \rightarrow \mathbb{R}^k$. 

As our algorithm depends on the linearity of the connection, we use the absolute value to define $\hat{b}(\theta,\pi)~=~|b(\theta, P(\theta,\pi))|$ as the loss barrier where $\pi \sim T$.\footnote{For a given two-layer feed-forward network with $h$ hidden neurons, $d$-dimensional input and $\theta_h$ representing its parameters, \citet{entezari2021role} showed that with probability $1-\delta$ over uniformly sampled $\theta_h$, there exists a permutation $\pi$ for which $\hat{b}(\theta_h,\pi) \leq \Tilde{O}(h^{-1/(2d+4)})$. \citet{stoica2023zipit} later obtained a tighter bound on the loss barrier by incorporating redundancy in the model.} Following~\citet{yang2021taxonomizing}, we compute the loss barrier at the mid-point by setting $\alpha=0.5$.\footnote{\textcolor{black}{Empirically, we found that the maximum loss barrier along the interpolation path occurs at $\alpha=0.5$ (see \autoref{fig:alpha} in Appendix).}} Given a model $\theta$, maximum number of permutations $n$ and threshold $\lambda$, our algorithm is as follows:
\begin{enumerate}[topsep=0.1cm, parsep=0pt, label=(\roman*)] %parsep=0pt, leftmargin=0.5cm, 
	\item Iteratively generate $n~(=|T|)$ random permutations of the base node $\theta$. 
	\item For each new node generated using $\pi_i \sim T$, check if there exists an edge between the current node and the base node by satisfying the linearity constraint, i.e., $\hat{b}(\theta, \pi_i)~\leq~\lambda$.
	\item Return the ratio $m(\theta,\lambda, n)$ between the number of edges in the graph ($e$) and the maximum number of possible edges ($e_\texttt{upper}$), i.e., $m(\theta,\lambda, n)~=~\frac{e}{e_\texttt{upper}}$.
	      % \end{equation}
\end{enumerate}

If the nodes are permutation invariant, the resulting loss barrier is small and the minima are considered \textit{connected} in the manifold.
\textcolor{black}{In other words, $m(\theta,\lambda, n)$ approximates a continuous manifold structure in the loss landscape where a high number of such connections, with $m(\theta,\lambda, n) \approx 1$, suggests the presence of a wide, relatively flat region of functionally equivalent minima — a \textit{manifold} in the sense of low-loss connectivity~\citep{simsek2021geometry}.}

\begin{figure}
% \begin{wrapfigure}{R}{0.475\textwidth}
	\begin{minipage}{0.5\textwidth}
		% \vspace{-10pt}
    \begin{algorithm}[H]
        \caption{Manifold detection: $m(\theta,\lambda, n)$}\label{alg:metric}
        \begin{algorithmic}
            \REQUIRE Model $\mathcal{M}_{\theta}$, Number of nodes $n$, Threshold $\lambda$, Loss function $L_D$ %Dataset $D$, 
            % \STATE $N \gets \{\}$
            \STATE $e \gets 0$, $\theta_1 \gets \theta$
            % \State $E \gets \{\}$
            % \State $l \gets L(\theta; x, y)$
            \FOR{$i~\text{in}~[1,n]$}
            \STATE Sample $\pi_i$ from $T$ without replacement 		    
            % \State Push $\pi_\theta$ to $N$
            \STATE $\theta_{i+1} \gets P(\textcolor{black}{\theta_i}, \pi_i)$  \hfill  $\triangleright$ Permuted model
            % \State $l' \gets L(\theta'; x, y)$
            \STATE $\hat{b}_i(\theta) \gets \lvert b(\theta, \theta_{i+1})\rvert$ \hfill $\triangleright$ \autoref{eq:barrier}
            \STATE \textbf{if}\phantom{0}{$\hat{b}_i(\theta) \leq \lambda$} \textbf{then}\phantom{0} $e \gets e+1$ % \hfill \Comment{Found an edge}
            \STATE $\theta_i \gets \theta_{i+1}$ %\hfill \Comment{To Ensure degree is 2 }
            \ENDFOR
            \STATE \textbf{return} $e/n$
        \end{algorithmic}
    \end{algorithm}
	\end{minipage}
    \hfill
    \begin{minipage}{0.4\textwidth}
      \centering
      \includegraphics[width=\linewidth]{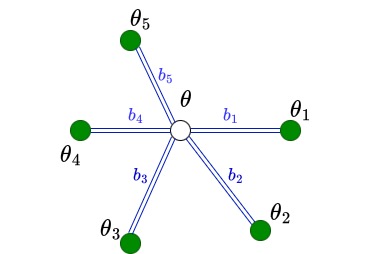}
      \captionof{figure}{\textcolor{black}{An example of constructed graph with $n=5$.}}
      \label{fig:diag}
    \end{minipage}%
% \end{wrapfigure}
\end{figure}

Unlike \citet{entezari2021role}, our goal is to compute the loss barrier between $\theta$ and $\theta_i$ that are generated from the same model, where $i \in [1, n]$.\footnote{\textcolor{black}{Our iterative generation of $\theta_i$ avoids selecting a \textit{bad} node—one that is not connected to any other nodes—which would reduce the number of valid comparisons when computing $m(\theta, \lambda, n)$. In such cases, the upper bound $e_\texttt{upper}$ would drop from $^nC_2$ to $[^nC_2 - (n - 1)]$, leading to wasted computation.}} \textcolor{black}{In this way, our iterative approach avoids computing barrier between all pairs of generated nodes which reduces algorithmic complexity from $O(n^2)$ to $O(n)$, since $e_\texttt{upper} = n$ matches the number of forward passes needed to compute the metric.} We provide pseudo-code in Algorithm \ref{alg:metric} \textcolor{black}{and an illustration in \autoref{fig:diag}}. We define our manifold metric for evaluating model expansion in the next section.

\subsection{Permutation Invariance and Expansion}

Let the parameters of the pre-trained base model be $\phi^*$. We perform a function-preserving expansion $\phi^*\rightarrow\theta$ using Net2net~\citep{chen2016net2net} to initialize new parameters. To ensure a fair evaluation of model expansion, we fix $n$ and $\lambda$ throughout the process. Here, $n$ determines the size of the graph, with higher values leading to a better estimate of the connectivity at the cost of more computations. On the other hand, $\lambda$ dictates the strictness of the connection between two minima to be linear, with smaller values enforcing stronger linearity constraints. The connectivity of a candidate model hence depends on the $\lambda$ associated with the base model. 
Accordingly, we define a set of loss barriers between permutations i.e., ${B}^* = \{\hat{b}(\phi^*, \pi_i);~~\forall~i \in [1, n]\}$ and define $\lambda = {B}^*_q$, i.e., the $q^{th}$ quantile of ${B}^*$ where $q \in (0, 1)$. We evaluate using the manifold metric $M$ defined as:
\begin{equation}\label{eq:metric}
	M(\theta, \phi^*) = m(\theta,{B}^*_q, n) - m(\phi^*, {B}^*_q, n) \\
    \approx m(\theta,{B}^*_q, n) - q
\end{equation}
\textcolor{black}{With $\lambda = B^*_q$, the proportion of permuted models of $\phi^*$ whose barriers are below $\lambda$ is approximately $q$ and hence we get $m(\phi^*, {B}^*_q, n)\approx q$.}
{Since $M(\theta, \phi^*)$ is a relative measure of connectivity rather than an absolute one, it ranks candidate models for expansion without being \textcolor{black}{over-dependent} on the high dimensionality of the models.} Although it primarily indicates changes in the geometric properties of the loss landscape, several other factors can influence its behavior. Curvature around the parameters $\phi^*$ matters since if the parameter lies on a sharp or noisy surface, the algorithm would find fewer edges, i.e., the optimal $q$ would be higher. Alternatively, if the base model is over-parameterized or contains dead neurons, ${B}^*_q$ will be smaller and the resulting metric may not indicate any change in the mode connectivity after the expansion, even if there is a performance gain.

\section{Experiments}\label{sec:experiments}

\subsection{Setup}

To evaluate our metric on model expansion, we first train a small base model with parameters $\phi$ until early stopping, at which point the parameters have evolved to $\phi^*$. We record the best validation performance as $A(\phi^*)$. The model is expanded to obtain a new set of parameters $\theta^{(0)}$ such that $A(\theta^{(0)}) = A(\phi^*)$, i.e. the expansion is function-preserving. We compute the manifold metric $M(\theta^{(0)}, \phi^*)$ using \autoref{eq:metric} (as a percentage). For simplicity, we refer to $M(\theta^{(t)}, \phi^*)$ as $M_t$ for the remainder of this work. The expanded model is trained for an additional $T$ epochs to obtain a performance gain $G_T = A(\theta^{(T)}) - A(\phi^*)$, where $\theta^{(t)}$ denotes the parameters after training the expanded model for $t\in[0, T]$ epochs. 
Although the first layer may appear specific, permuting these neurons can lead to a substantial shift in the parameter location, which is crucial for determining the geometry of the loss landscape~\citep{JMLR:v23:20-069}.

We run our experiments with CNNs on CIFAR10/100, ResNet18~\citep{he2016deep} on CIFAR10, and TinyBERT/MiniBERT~\citep{BERT} on WikiText-103~\citep{merity2016pointer}. Unless specified, hyper-parameters were determined with a grid search, detailed in Appendix~\ref{app:hyper}.
CNNs/ResNets are expanded along the width/channels, and BERT models along the width of the intermediate layer in the feed-forward network. 
{We use cross-entropy loss for all our experiments. Validation accuracy for predicting class probabilities in CIFAR10/100 experiments and validation loss for predicting the original token in the masked language modeling task for WikiText-103 experiments were chosen as the performance evaluation criteria $A(\cdot)$.}
Permutations are also conducted along these specific axes (Appendix~\ref{app:permute}). All results are averaged across $10$ seeds and presented along with the standard deviation.

\subsection{Comparison with Zero-Cost Baselines}

{Our proposed metric reduces training costs as it does not require training the candidate models to predict their final performance. This makes it a cost-efficient zero-cost proxy for ranking candidate expansion models.} We therefore start by comparing the manifold metric with several existing zero-cost proxy baselines such as gradient norm (GradNorm)~\citep{pmlr-v80-chen18a}, Jacobian covariance (Jacov)~\citep{mellor2021neural}, and pruning-based methods like single-shot network pruning (SNIP)~\citep{lee2018snip}, gradient signal preservation (Grasp)~\citep{wang2020picking}, and synaptic flow (SynFlow)~\citep{tanaka2020pruning} that have previously been adapted as zero-cost proxies by~\citet{abdelfattah2021zero}. Among these baselines, Jacov and SynFlow have shown a high correlation with performance~\citep{white2021powerful}. \textcolor{black}{Here, correlations are computed over 10 different base models (random seeds), each leading into multiple candidate expansion models.} We ensure that our metric and the baselines use a fixed number of training examples i.e., a single batch of the training data for their computation.

\begin{figure*}[thb!]
    \centering
    \includegraphics[width=0.9\textwidth]{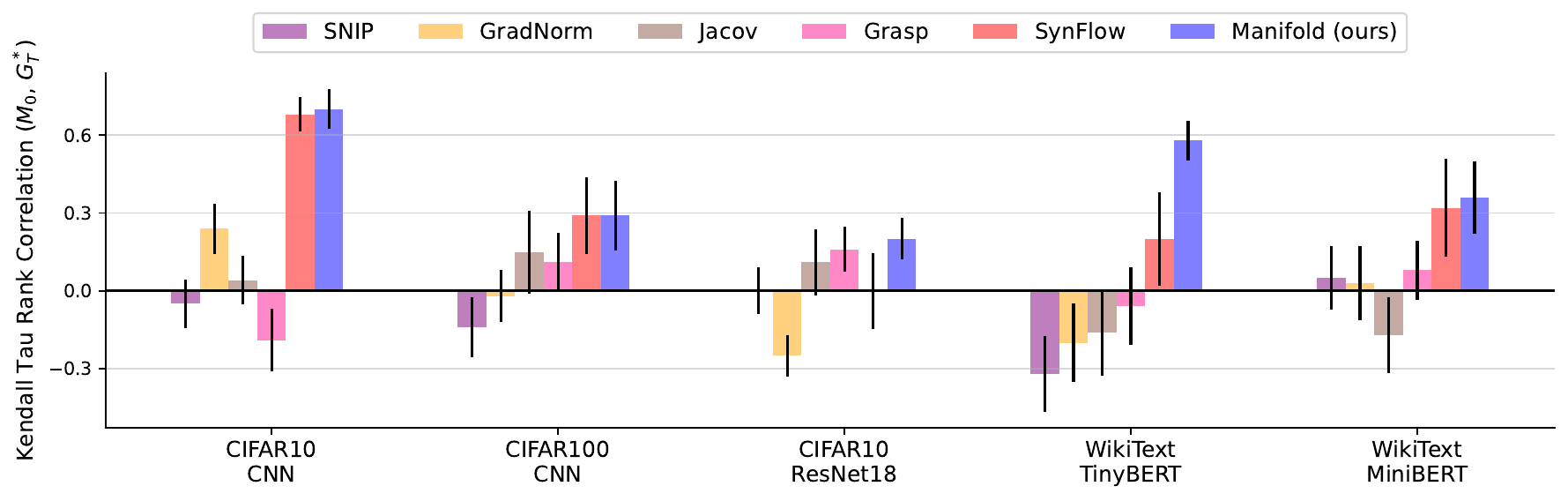}
    \caption{Comparison between our manifold metric and existing zero-cost proxies in terms of Kendall Tau rank correlation with the performance gain $G^*_T$. The horizontal axis indicates the dataset and the base model. Correlation with our metric remains consistently positive and competitive to the second-best baseline, whereas other baselines fluctuate between positive and negative.}
    \label{fig:all_tau}
\end{figure*}

\autoref{fig:all_tau} shows the correlation between the metrics and the highest performance gain $G^*_T$ observed after expanding pre-trained models and training them for at least $T$ epochs. We use Kendall Tau rank correlation~\citep{Puka2011} throughout our experiments following~\citet{white2021powerful}\footnote{Similar results using other correlation metrics are presented in Appendix~\ref{app:corr_met}.}. The manifold metric is the \textit{only} metric that remains positively correlated for all settings, while the selected baselines can vary, making them unreliable as general-purpose metrics across the chosen benchmarks. This showcases the consistency and reliability of manifold metrics in ranking candidate models for expansion without training, as it remains best or competitive to the second-best baseline. Meanwhile, SynFlow shows a high correlation with performance in 3 out of 5 settings and Jacov shows a low correlation that remains close to $0$, suggesting that unlike in NAS, their performance remains poor and inconsistent when evaluated on pre-trained model expansion.

A detailed comparison between the manifold metric and baselines in terms of computation time is presented in Appendix \ref{app:baselines}. We also provide a sensitivity analysis in Appendix~\ref{app:sens} that shows a trend in the correlation obtained by varying $n$ and $q$. In almost all cases, increasing $n$ improves the overall representative power of the metric for estimating the change in manifold. Moreover, we find that it is possible to obtain an optimal value for the threshold quantile $q$ without running an exhaustive grid search.

\subsection{Image Classification}\label{sec:image}

To analyze changes in the loss landscape when training an expanded model, we compare our manifold metric with the sum of training losses per epoch (SoTL-E)~\citep{ru2021speedy}, which has been shown to have a high correlation with final validation accuracy in NAS literature~\citep{white2021powerful}. In the following experiments, we track the performance gain $G_t$ and the manifold metric $M_t$ for $(q, n)=(0.4, 1000)$. We also present a brief sensitivity analysis in terms of correlation between $M_t \%$ and $G^*_T$ for $q \in \{0.1, 0.2, 0.4\}$.

\begin{figure*}[thb!]
    \centering
    \resizebox{\linewidth}{!}{
        \includegraphics[width=0.66\linewidth]{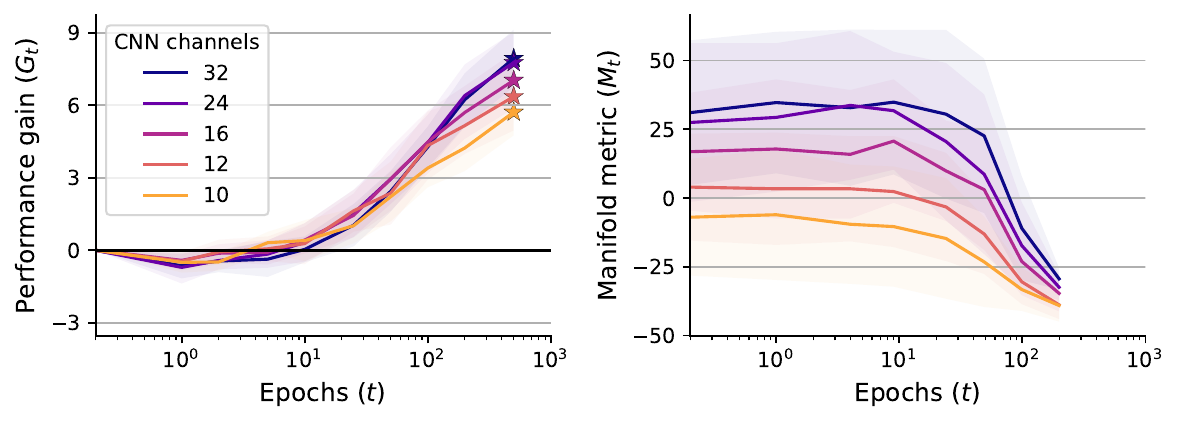} 
    	\includegraphics[width=0.32\linewidth]{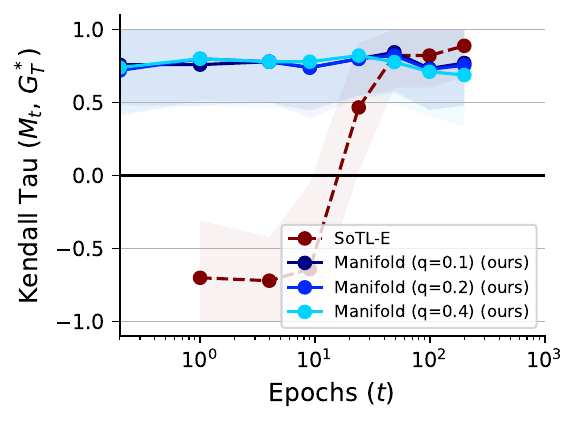}
    }
    \resizebox{\linewidth}{!}{
        \includegraphics[width=0.66\linewidth]{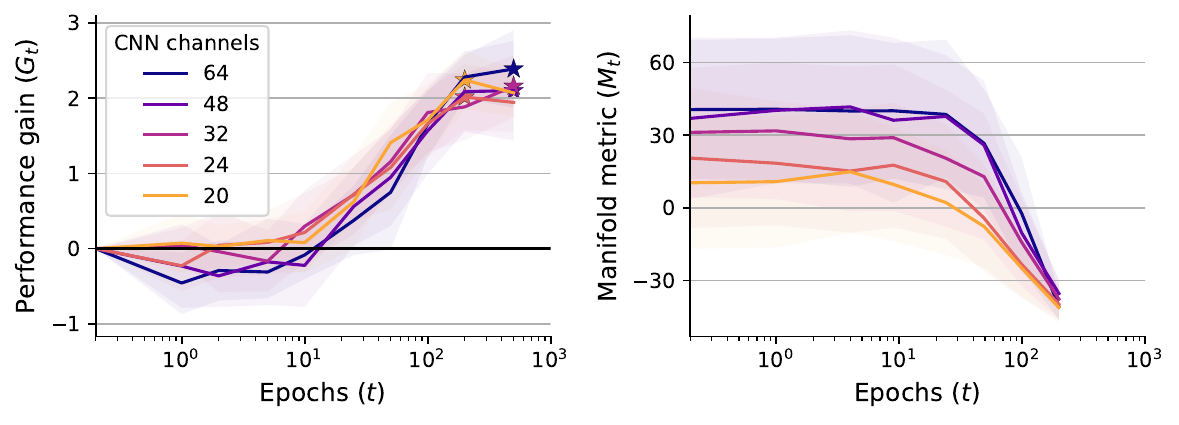}
    	\includegraphics[width=0.32\linewidth]{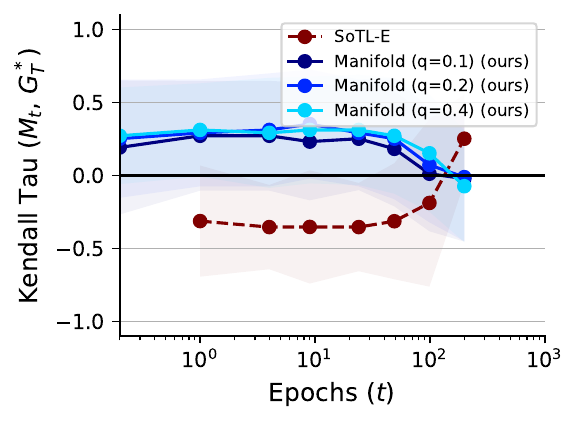}
    }
    \caption{Performance gain $G_t$ (first column; $\star$ denotes the best performance gain $G^*_t$), manifold metric $M_t \%$ (second column) and correlation between $M_t$ and the highest gain $G^*_T$ (third column) on expanding and training CNN on CIFAR10 (top) or CIFAR100 (bottom) for $T=500$ epochs. Overall, $M_t$ remains constant and proportional to $G^*_T$ from the beginning and drops during the later phase of training. Comparison of correlations shows that SoTL-E is correlated to the gain $G_t$ at a given epoch $t$ and therefore results in a negative correlation with $G^*_T$ initially. $M_t$ positively correlates early on, making it more reliable for comparing expanded models.} 
    \label{fig:image_manifold}
\end{figure*}

\autoref{fig:image_manifold} reveals that when training expanded CNNs on CIFAR10 (first row), increasing the number of channels results in larger performance gain (first column), and the manifold metric is positively correlated (second and third columns). We also observe that $q=0.4$ performs marginally better than lower values of $q$. On CIFAR100 (second row), over-fitting occurs at the end of training and $G_t$ improves marginally with an increasing number of channels, resulting in an overlap of overall performance gains. As a result, the correlation is initially small but positive. 

For both datasets, $M_t$ remains constant in the beginning (up to $t\approx 20$) and drops during the later phase of training. This observation indicates that nodes generated by randomly permuting $\theta^{(t)}$ for $t\leq 20$ can be connected via linear paths more often. But when the model is trained further, the parameters enter a sharper region in the loss landscape with higher barrier sizes and therefore result in smaller $m(\theta^{(t)},\lambda, n)$ in \autoref{eq:metric}. For instance, in CIFAR100 for $q=0.4$ (and therefore $m(\phi^*,\lambda, n)\approx 40$), $m(\theta^{(t)},\lambda, n)$ approaches $0$ when over-fitting occurs for $t\geq 200$ and, as a result, we observe $M_t\approx -40\%$.

{Additionally, after a certain level of expansion, if the performance plateaus, the manifold metric still correlates better with performance than other baselines. This phenomenon is observed in \autoref{fig:image_manifold_ablation} (first row), where candidate models with 24 and 32 channels perform almost identically (first column), unlike others. A similar pattern is observed in the manifold metric plot (second column), where there’s a clear gap between the metrics for models with 10-16 channels, but this gap narrows between 24 and 32 channels, indicating that further expansion is unlikely to result in any significant improvement in performance.}

We also observe that SoTL-E initially results in a highly negative correlation, which suggests SoTL-E can correlate well to the performance gain at a given epoch $G_t$ rather than the final performance gain $G^*_T$. While SoTL-E is considered an oracle scoring method, its effectiveness appears to be highly dependent on the current parameter location, which is undesirable as these baselines could require training the candidate models of expansion for extended (high query) time before accurately ranking them~\citep{white2023neural}. Therefore, even when we train the expanded model, the manifold metric can provide an approximate rank for generalization performance earlier on.

\subsubsection{Ablations on CNN Expansion}\label{sec:ablation}
In the previous results, we expanded the base CNN model by adding more channels in its first layer. To highlight our method's robustness to the expansion being performed, we provide an ablation study where we \textbf{(i)} expand width in all CNN layers on CIFAR100, \textbf{(ii)} increase depth of the CNN model on CIFAR100, and \textbf{(iii)} expand a ResNet18 model~\citep{he2016deep} by width on CIFAR10. We use the same CNN model from the previous section for CIFAR100 ablations. We also use Net2net~\citep{chen2016net2net} for initializing new parameters so that the expansion is function-preserving. We set $q=0.4$ and $n=1000$ in all three setups.

\begin{figure*}[thb!]
    \centering
    \resizebox{\linewidth}{!}{
        \includegraphics[width=0.32\linewidth]{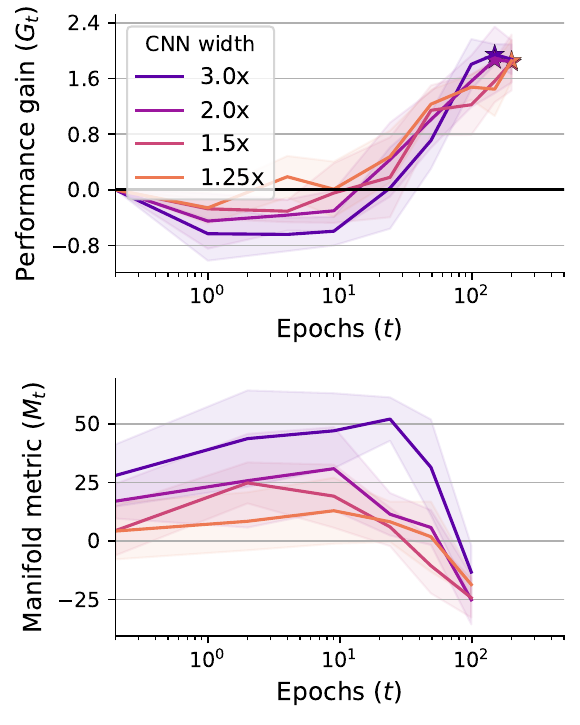} 
        \hfill
        \includegraphics[width=0.32\linewidth]{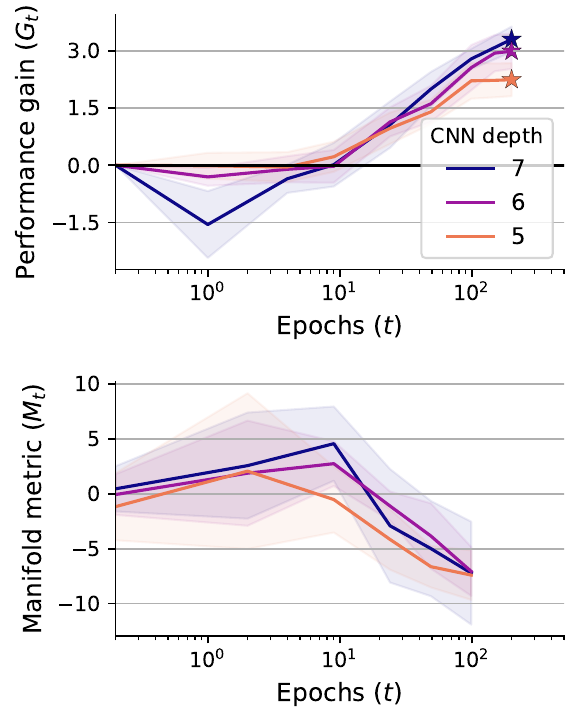}
        \hfill
        \includegraphics[width=0.32\linewidth]{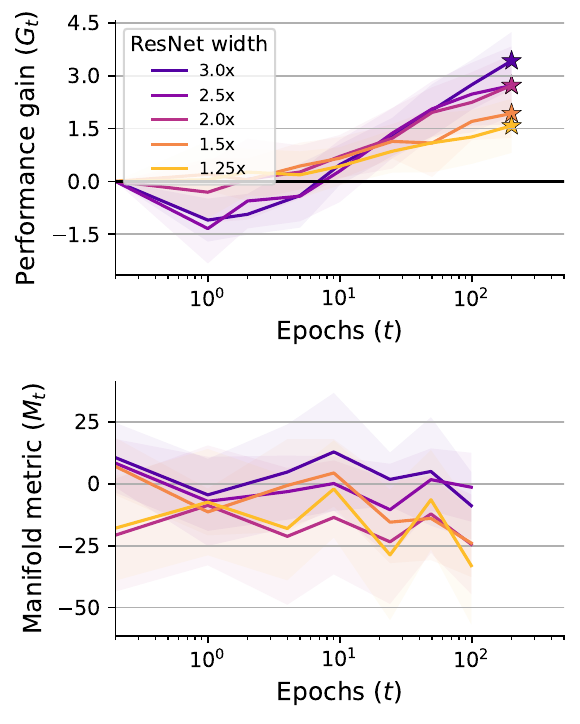} 
    }
    \caption{Performance gain $G_t$ (first row; $\star$ denotes the best gain $G^*_t$) and manifold metric $M_t \%$ (second row) for CNN width expansion on all layers (first column) and depth expansion on CIFAR100 (second column), and ResNet width expansion on CIFAR10 (third column). $M_t$ positively correlates with the level of expansion.} 
    \label{fig:image_manifold_ablation}
\end{figure*}

When increasing the width of all the convolutional layers, \autoref{fig:image_manifold_ablation} (first column) shows the performance gain to be similar for all expansion factors, but $M_t$ attains an overall positive correlation early on during training. We also observe that when the number of channels is increased up to $24$, the increase in $M_t$ is more pronounced. Interestingly, $M_t$ increases in the first few epochs, instead of remaining constant, before decreasing for later epochs. During depth expansion (second column), adding more layers results in increasing performance. While the manifold metric correlates with $G^*_T$ on average, there is significant overlap across different levels of expansion. When expanding the convolutional layers in a ResNet18 model (third column) by width, while there is a variance overlap in $M_t$, it remains positively correlated to $G^*_T$.

Next, we evaluate our metric on various models with similar architectures. For this experiment, we expand a base CNN architecture by increasing its width and depth while keeping the number of parameters relatively similar to the base model. The goal of this ablation study is to assess the robustness of our metric in predicting performance towards different types of expansion. We generate ten such candidate models (details provided in Appendix \ref{app:hyper}) and train them until convergence using AdamW.

\begin{wrapfigure}{r}{0.5\textwidth}
 % \vspace{-15px}
\begin{center}
% \begin{figure}
    \centering
    \includegraphics[width=\linewidth]{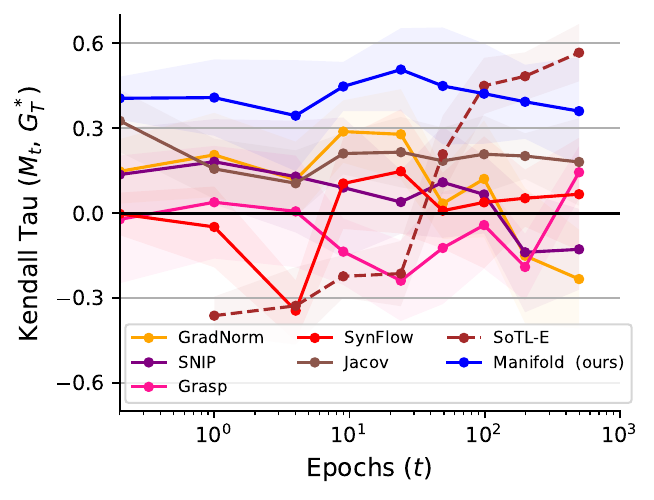} 
    \caption{Comparing rank correlation between different metrics and the highest gain $G^*_T$ observed after expanding (by both width and depth) and training CNN models on CIFAR10. Overall, our manifold metric correlates better with $G^*_T$ throughout training, making it more reliable for comparing expanded models. All other baselines have poor correlation, except Jacov, which has the second-best correlation after expansion.}
    \label{fig:same_size}
% \end{figure}
\end{center}
\vspace{-25px}
\end{wrapfigure}

In \autoref{fig:same_size}, we plot Kendall Tau rank correlation between manifold metric $M_t$ and the highest performance gain $G^*_T$ on expanding and training the candidate models for $T=500$ epochs. Overall, $M_t$ exhibits stronger correlation with $G^*_T$ throughout training, making it more reliable for comparing different types of models. On the other hand, the zero-cost baselines exhibit poor correlation throughout training except Jacov which has second-best correlation just after expansion. SoTL-E exhibits similar behavior, as previously observed in \autoref{fig:image_manifold}, where it initially has a negative correlation with $G^*_T$. We observe similar correlation trends using Adam optimizer (see Appendix \ref{app:adam}). These results show that our metric ranks models based on their performance, not just their size, as shown by its superior performance compared to existing baselines.

\subsection{Language Modeling}\label{sec:BERT_language}

While the existence of LMC in the loss surfaces of MLPs and CNNs has been broadly explored, understanding of the loss landscape geometry in Transformers remains limited~\citep{yang2021taxonomizing}. Nevertheless, \citet{qin2022exploring} suggested that LMC emerges during fine-tuning, where a pre-trained model is evaluated on different tasks, motivating us to study the manifold metric with encoder-only Tiny/MiniBERT models, specifically with masked language modeling on WikiText-103.
We train a base model for $5$ epochs before expansion, then expand the model by increasing the intermediate layer width of its feed-forward networks. The expanded model is further trained for $15$ epochs, with the performance measured as a negative log of validation loss.

\begin{figure*}[thb!] 
    \centering
    \resizebox{0.9\linewidth}{!}{
        \includegraphics[width=0.66\linewidth]{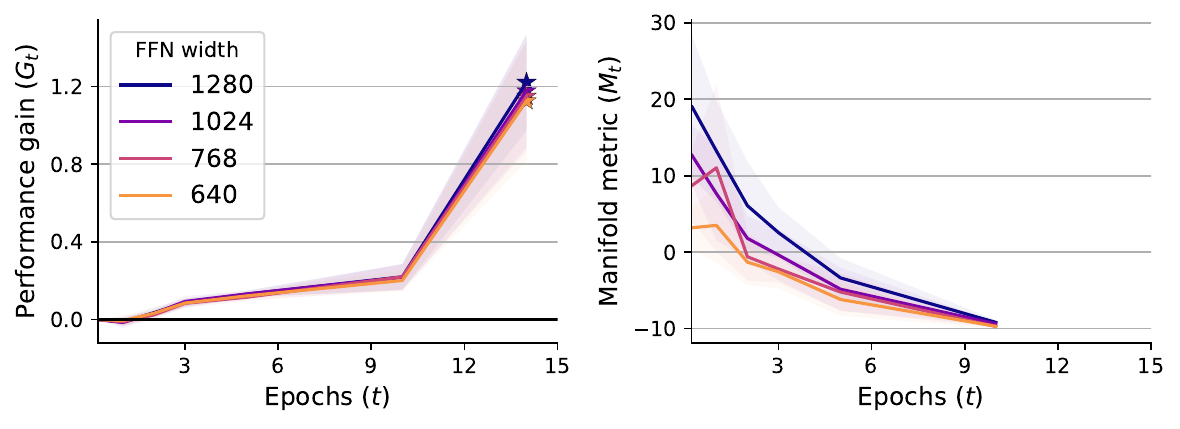}
        \includegraphics[width=0.32\linewidth]{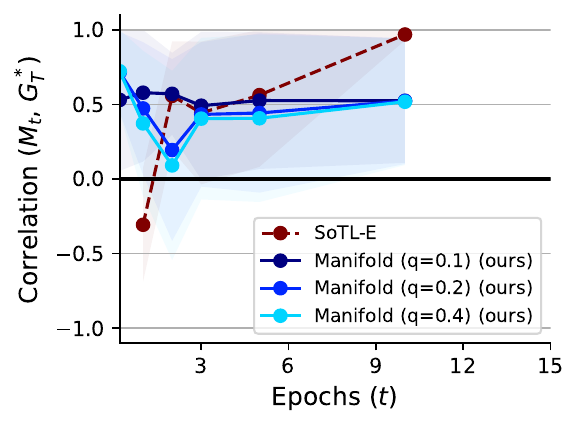}
    }
    \resizebox{0.9\linewidth}{!}{
        \includegraphics[width=0.66\linewidth]{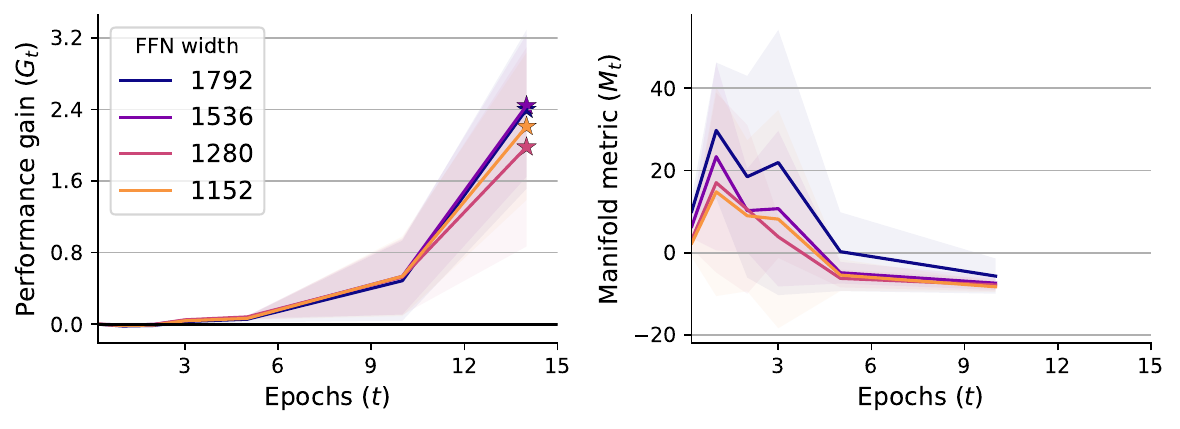}
        \includegraphics[width=0.32\linewidth]{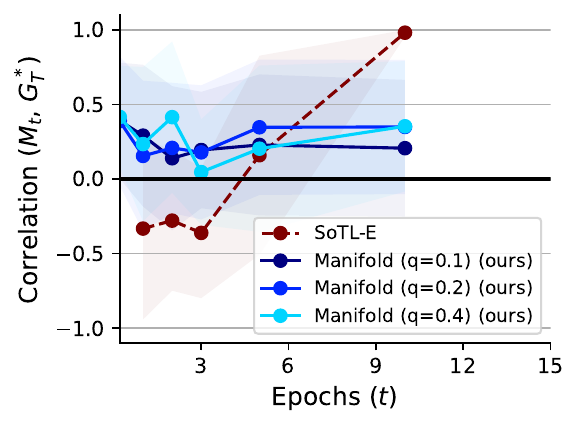}
    }
    \caption{
         Performance gain $G_t$ (first column; $\star$ denotes the best gain $G^*_t$), manifold metric $M_t \%$ (second column) and correlation between $M_t$ and $G^*_T$ (third column) for FFN expansion in TinyBERT (first row) and MiniBERT (second row) on WikiText-103 dataset. The manifold metric $M_t$ drops as the expanded model is trained, but remains positively correlated with $G^*_T$. However, SoTL-E shows a negative correlation at the start of training. 
     }
     \label{fig:BERT_manifold}
\end{figure*}

\autoref{fig:BERT_manifold} (first column) shows that increasing the FFN width in both TinyBERT and MiniBERT contributes to performance gain. However, there is an initial increase in the manifold metric followed by a significant drop as the model is trained after expansion (second column), which is interesting. While several plausible reasons for this exist, we include two ablations in Appendix~\ref{app:transformer} to verify whether this phenomenon specifically occurs because of the expansion: we \textbf{(i)} train a Transformer from a random initialization and \textbf{(ii)} permute self-attention weights instead of FFN weights. Both cases observe a similar drop in the manifold metric, suggesting a link with the overall training dynamics in the Transformer and language modeling tasks.
Despite the drop, a positive correlation still exists throughout training after expansion (third column) for both BERT models. This is crucial since SoTL-E begins with a negative correlation and takes at least $2$ epochs to show a positive correlation.

\section{Discussion and Limitations}

Our manifold metric quantifies geometric properties of the loss landscape beyond local measures like curvature or sharpness by spanning multiple minima. This allows it to  cover a broader region, including the barriers between minima. The metric provides explicit control over the spanned area through a hyperparameter $n$. While inspired by LMC, our approach does not impose strict linearity. Instead, it uses a threshold-based hyperparameter $\lambda$  to capture nonlinearity in the loss surface. Although establishing  clear causal relationships would require further investigation~\citep{dziugaite2020search}, mode connectivity is often associated with better generalization~\citep{mirzadeh2020linear, neyshabur2020being}. Motivated by this, we used mode connectivity to compare candidate expansion models, which consistently outperformed Hessian-based baseline such as Grasp.

Throughout this work, we observed that model expansion can lead to an overlap in performance gain across seeds for different expansion factors. For example, in CNN width expansion on CIFAR100, the difference between $G^*_T$ for different expansion factors was almost similar. Despite this, $M_t$ continues to consistently exhibit a positive and a higher correlation with $G^*_T$ as compared to all other baselines. Therefore, even when there is significant variance in performance or correlation, the metric is sufficient to make a personalized choice on whether to continue expansion or not.

Our experiments mainly focused on expanding small-scale models such as CNNs and TinyBERT. However, our results indicate that the manifold metric is more reliable compared to other baselines as both model and dataset complexity increase, serving as a promising starting point. Our sensitivity analysis on these setups also suggests a trend that leads to better correlation as we increase the number of permutations and threshold quantiles, allowing one to avoid running an exhaustive grid search over these hyperparameters potentially. While similar experiments could be conducted on large-scale models using alternative techniques, {we argue that such setups with large base models would exhibit weaker correlations and standard deviations since they already perform near optimally before expansion. We also observe this phenomenon in \autoref{fig:image_manifold} and \autoref{fig:BERT_manifold} where there is overlap in the final performance gains of the different candidate models for the bigger base models across seeds.}

{Finally, we would like to point out that while} different initialization methods exist for the newly added parameters after expansion, we use Net2net~\citep{chen2016net2net} in our experiments. Since the manifold metric requires the parameters to be located at a minimum in the loss landscape, exploring different initialization methods is orthogonal to our experiments as long as the function is preserved after expansion.

\section{Conclusion}
% \vspace{-0.1in}
In this paper, we observe the effects of expanding a model through the perspective of the loss landscape. We introduce a metric that estimates the size of the minima manifold and shows that it can reliably evaluate the benefits of model expansion. Our results indicate that this metric can correctly rank models for image classification tasks, unlike other existing baselines, without requiring training the model for several epochs.

Experiments with Transformers suggest that expansion can lead to a more complex and non-linear loss landscape, which comes with a significant drop in manifold metric that underscores the limited understanding of their training dynamics. The applications of the proposed metric are not limited to architecture search. It evaluates whether any change in the loss landscape results in an increase in mode connectivity and since changes in data distributions also result in changes in the loss landscape, we believe it helps evaluate whether a given model can perform well under distribution shifts. Therefore, we believe that such metrics can be useful for continual learning, online learning, and out-of-distribution generalization, among others. \textcolor{black}{While we provide some promising preliminary results from online learning experiments in Appendix \ref{app:online}, we hope that these directions can be explored in future work at larger scale.}

\subsubsection*{Acknowledgements}

Pranshu Malviya is partially supported by the Merit scholarship program for foreign students (PBEEE) by Fonds de Recherche du Qu\'{e}bec Nature et technologies (FRQNT). Jerry Huang is partially supported by a Natural Sciences and Engineering Research Council of Canada (NSERC) Canada Graduate Scholarship, FRQNT training scholarship and Hydro-Qu\'{e}bec Excellence Scholarship. Sarath Chandar is supported by the Canada CIFAR AI Chairs program, the Canada Research Chair in Lifelong Machine Learning, and the NSERC Discovery Grant. This research was enabled in part by compute resources provided by Mila (\href{https://mila.quebec/}{mila.quebec}) and the Digital Research Alliance of Canada (\href{https://alliancecan.ca/}{alliancecan.ca}).

\bibliography{refs}
\bibliographystyle{collas2025_conference}

\appendix
\section*{Appendix}
\section{Impact Statement}
Large neural networks require substantial computational resources to be trained from scratch, and computing is not only finite but also expensive. Expanding smaller pre-trained models has the potential to make deep learning research quicker, more affordable, and more accessible to groups with limited resources. Furthermore, the large-scale infrastructures often required for training large models emit considerable amounts of carbon dioxide (CO2). However, it is challenging to determine whether expansion would have a positive impact on the environment as, while it would help reduce the carbon footprint of each model, it would also allow training more models. Overall, we believe expansion is a step toward the re-usability of neural networks and, more generally, more sustainable deep learning. 

\section{Implementation Details and Additional Analysis}\label{app:implemetation}

In this section, we provide the implementation details and results that were not included in the main content. In section \ref{app:permute}, we describe the process of generating permutation used in our experiments. In section \ref{app:hyper}, we provide details about the datasets and models along with corresponding hyper-parameters values. Next, we provide a sensitivity analysis in section \ref{app:sens} and a comparison between our method and the zero-cost baselines using different correlation metrics in section \ref{app:corr_met}. This is followed by additional experiments for analyzing the manifold size in an image classification task in section \ref{sec:image} and language modeling tasks in section \ref{app:transformer}. We use the Jax library~\citep{jax2018github} in Python for our implementation. Experiments were conducted on NVIDIA V100 or A100 GPU machines with 32 GB and 40 GB memory, respectively.

\subsection{Generating Random Permutations}\label{app:permute}

As mentioned in section \ref{sec:experiments}, we perform permutations on the neurons or channels of the first hidden layer in image classification tasks and the intermediate layer neurons of the first encode layer in the BERT models. For a given model, let $\theta$ represent the parameters located at a minimum. If the width of the first layer is $l$ ($\geq 4$), we denote the corresponding input and output parameters of the first layer as the matrix $\theta^{i}_1$ with a shape of $x\times l$ and the matrix $\theta^{o}_1$ with a shape of $l\times y$. To permute the neurons in the layer in a function-preserving manner, we shuffle the columns of $\theta^{i}_1$ and the rows of $\theta^{o}_1$ using the same criteria. Since there are $l$ neurons, there are $l!$ possible permutations that can be applied to obtain the same function.

The proposed metric requires finding edges through mode connectivity between minima. One might argue that swapping only two neurons in the layer can result in a closely located pair of minima. While this holds for smaller pre-trained models, employing this approach for larger base models may lead to a sample that only contains dead neurons. As a result, $B^*_q$ can become very small, reducing the representative power of the manifold metric, as discussed in section \ref{sec:methodology}. On the other hand, using entirely random permutations may result in generated minima located very far from each other. This is again not desirable since it will result in high loss barriers. 

\begin{figure}[htb!]
    \centering
    \includegraphics[width=0.7\linewidth]{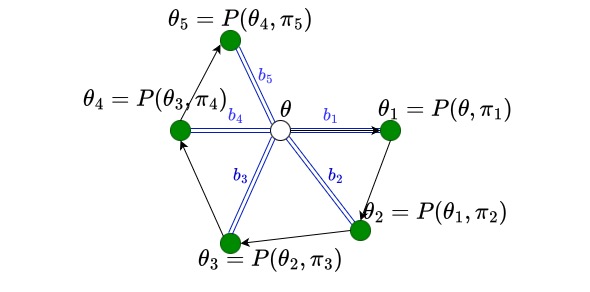} 
    \caption{\textcolor{black}{An example of our node generating approach with $n=5$. Starting from $\theta$, we apply permutation $\pi_i$ iteratively on previous node to generate a new node.}}
    \label{fig:diag2}
\end{figure}

\begin{figure*}[htb!]
    \centering
    \resizebox{\linewidth}{!}{
        \includegraphics[width=0.52\linewidth]{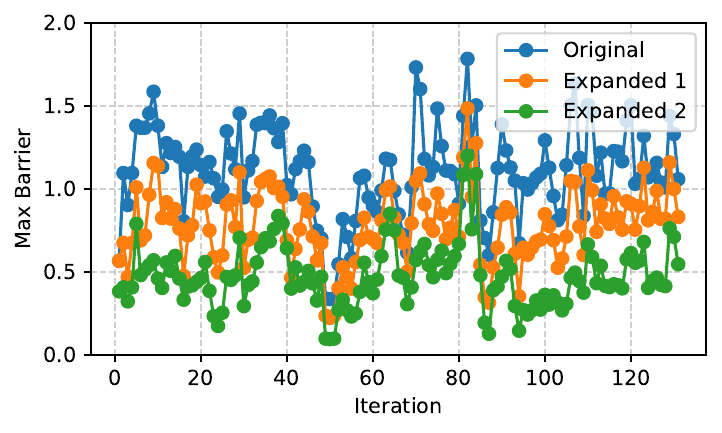} 
    	\includegraphics[width=0.4
\linewidth]{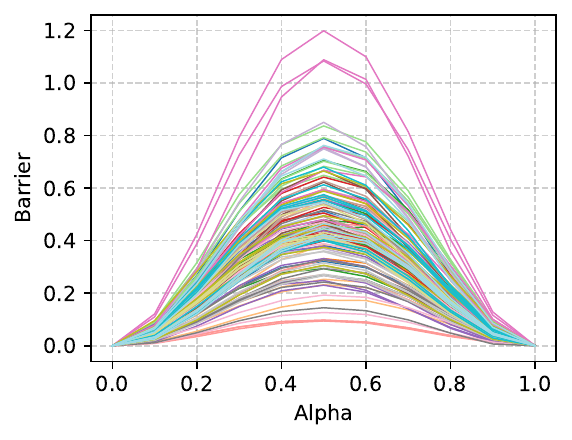}
    }
    \caption{\textcolor{black}{(i) Evolution of maximum loss barrier $b_i$ obtained between base node and a $n$ randomly generated nodes using Algorithm \ref{alg:metric} for Original/base and the two expanded models (candidate 1 and candidate 2) from \autoref{tab:same_size}. We observe that our sampling strategy approximate uniformity as $n$ increases. (ii) Comparing linear interpolation obtained for \textit{Expanded 2}. We observe that the maximum loss barrier along the interpolation path occurs at $\alpha = 0.5$.}}
    \label{fig:alpha}
\end{figure*}

Therefore, as a trade-off between the distances between minima and the representative power of the metric, we iteratively sample $\lceil\log_2 l\rceil$ pairs of neurons and swap their position with replacement to obtain a new permutation. This way, the iteratively generated minima do not lie too far from one another in the loss landscape, making it easier to find edges between them. \textcolor{black}{Based on the example shown in \autoref{fig:diag}, we illustrate our node generating approach in \autoref{fig:diag2}. We also plot evolution $b_i$ in \autoref{fig:alpha} which suggests that our sampling strategy approximate uniformity as $n$ increases. }

\subsection{Datasets and Models}\label{app:hyper}

In \autoref{tab:dataset}, we provide a summary of all datasets used in our experiments.

\begin{table*}[htb!]\caption{Dataset details}\label{tab:dataset}
\centering
\begin{tabular}{@{}cccc@{}}
\toprule
Dataset & Train Set & Validation Set & Input shape \\ \midrule
CIFAR10  & 40K & 10K & MLP: (3072); CNN: (32,32,3)\\
CIFAR100 & 40K & 10K & (32, 32, 3) \\
WikiText-103  & 1.8M & 3.76K & embedding dimension=$128$, ~sequence length=$512$ \\ \bottomrule
\end{tabular}
\end{table*}

The tasks consist of:
\begin{itemize}
    \item CIFAR10 is another multiclass image classification task whose training and validation sets consist of RGB images that are separated into $10$ distinct classes, with each class containing an equal number of samples across the training and validation sets. 
    \item CIFAR100 is similar to CIFAR10, except that images are now separated into $100$ classes with an equal number of samples within each class.
    \item WikiText-103 is a collection of over $100$ million tokens extracted from the set of verified Good and Featured articles on Wikipedia. Since different articles, i.e., training samples, may have different lengths, the explicit size of the training and validation splits does not correspond to the number of training examples. Rather, this can be roughly estimated as the total number of tokens divided by the maximum sequence length, which is $512$.
\end{itemize}

In \autoref{tab:models_image}, we provide the details about deep learning models and the levels of expansion performed in them. $F(.)$ stands for the fully connected layer with a number of units in between parathesis. $C(.)$ represents convolutional layers with the number of channels in between parathesis with a fixed kernel size of $3\times 3$. We expand the first convolutional layer in both CNNs used in our experiments in \autoref{fig:image_manifold} .

\begin{table*}[!htb]
\caption{Base models and expansion details for image classification tasks. }\label{tab:models_image}
\centering
\resizebox{\textwidth}{!}{
\begin{tabular}{@{}ccc@{}}
\toprule
Model           & Architecture & Expanded layers in candidate models\\ \midrule
CNN (CIFAR10)       &    $ C_1(8) - C_2(32) - MaxPool(2) - F_1(256)$  & $\{C_1(10),~C_1(12),~C_1(16),~ C_1(24),~ C_1(32)\}$     \\
CNN (CIFAR100)      &  $  C_1(16) - C_2(64) - C_3(64)- C_4(64) - MaxPool(2) - F_1(256) $    & $\{C_1(20),~C_1(24),~C_1(32),~C_1(48),~ C_1(64)\}$       \\
\bottomrule
\end{tabular}
}
\end{table*}

For the experiments in \autoref{sec:ablation}, we expand (i) all the convolutional layers of CNN (CIFAR100) by width each with a given factor $\in \{1.25, 1.5, 2.0, 3.0\}$ and (ii) Add convolutional layers between $C_3(64)$ and $C_4(64)$. For the ResNet18 expansion, we increase the width of each convolutional layer present in the base model by a given factor $\in \{1.25, 1.5, 2.0, 2.5, 3.0\}$. We provide the details of each expanded model maintaining similar number of parameters in \autoref{tab:same_size}.

\begin{table}[h]
    \centering
    \caption{Candidate model details for ablation study in \autoref{sec:ablation} involving expansion with similar number of parameters.}
    \begin{tabular}{ll}
        \toprule
        \textbf{Model} & \textbf{Configuration} \\
        \midrule
        base  & $C_1(8) - C_2(32) - MaxPool(2) - F_1(256)$ \\
        candidate\_1  & $C_1(16) - C_2(32) - MaxPool(2) - F_1(256)$ \\
        candidate\_2  & $C_1(32) - C_2(32) - MaxPool(2) - F_1(256)$ \\
        candidate\_3  & $C_1(8) - C_2(8) - C_3(32) - MaxPool(2) - F_1(976)$ \\
        candidate\_4  & $C_1(16) - C_2(16) - C_3(32) - MaxPool(2) - F_1(976)$ \\
        candidate\_5  & $C_1(8) - C_2(32) - C_3(32) - MaxPool(2) - F_1(976)$ \\
        candidate\_6  & $C_1(8) - C_2(8) - C_3(8) - C_4(32) - MaxPool(2) - F_1(3712)$ \\
        candidate\_7  & $C_1(8) - C_2(8) - C_3(16) - C_4(32) - MaxPool(2) - F_1(3712)$ \\
        candidate\_8  & $C_1(8) - C_2(16) - C_3(16) - C_4(32) - MaxPool(2) - F_1(3712)$ \\
        candidate\_9  & $C_1(16) - C_2(16) - C_3(32) - C_4(32) - MaxPool(2) - F_1(3584)$ \\
        candidate\_10 & $C_1(32) - C_2(32) - C_3(32) - C_4(32) - MaxPool(2) - F_1(3584)$ \\
        \bottomrule
    \end{tabular}

    \label{tab:same_size}
\end{table}

We followed the openly available MaskedLM BERT configuration in the language modeling experiments described in \autoref{sec:BERT_language}~\citep{wolf2019huggingface}. The specific details of TinyBERT and MiniBERT are provided in \autoref{tab:models_BERT}. We perform expansion along the intermediate width of the FFN part in each encoder layer of the models. The widths in the candidate models are given in the final row in the table.

\begin{table*}[!htb]
    \caption{Base models for language modeling tasks and expansion details. }\label{tab:models_BERT}
    \centering
    \resizebox{\linewidth}{!}{
        \begin{tabular}{@{}ccc@{}}
        \toprule
            & TinyBERT & MiniBERT \\ \midrule
        Number of attention heads & $2$         & $4$         \\
        Number of encoder layers          & $2$         & $4$         \\
        Hidden dimension          & $128$       & $256$       \\
        Intermediate width       & $512$       & $1024$      \\ \midrule
        Intermediate widths in expanded models & $\{640,  768, 1024, 1280\}$ & $\{1152, 1280, 1536, 1792\}$ \\
        \bottomrule
        \end{tabular}
    }
\end{table*}

For the image classification task results shown in \autoref{fig:all_tau}, we used AdamW optimizer with a learning rate of $0.001$, weight decay of $0.0001$, and beta values ($0.9$, $0.999$) to obtain the pre-trained models. The batch size was fixed to $512$ for CIFAR10 and CIFAR100. For language modeling tasks, we used Adam optimizer with the learning rate $0.0003$ and beta values ($0.9$, $0.999$). We use a cosine learning rate scheduler with a warmup of $1000$ steps. The batch size was fixed at $8$ per GPU core for training. We compute the manifold metric and zero-cost baselines on a single batch of the training dataset and show the zero-cost correlation obtained for the metrics across all the tasks. 

For the rest of the experiments described in section \ref{sec:image} and section \ref{sec:BERT_language}, we fixed the pre-trained base model $\phi^*$ obtained using early-stopping criteria. We perform the expansion as described in the previous section and train the model to analyze how manifold metric evolves. This is done by storing checkpoints at multiple epochs and computing the metrics across them.

We show the manifold metric results obtained from the best-performing hyper-parameter setup by performing a grid search over different threshold quantiles $q$ and the number of nodes $n$. For grid search over $q$, we use the set $Q=\{0.05, 0.1, 0.2, 0.4\}$, and for $n$, we use the set $N=\{50, 100, 250, 500, 1000\}$. Note that, to reduce computation time, we do not run the algorithm \ref{alg:metric} separately for each combination $(q,n)$. Instead, we obtain the set $B^*=\{\hat{b}_i(\phi^*):~i\in [1,\max_\eta N]\}$, and compute the $B^*_q$ from the subset $B^*[1:n]$ for $q\in Q$ and $n\in N$ where $[:]$ represents slice of the set given the indices. Therefore, the computation requirement for hyper-parameter search is proportional to the maximum value in $N$.

\subsubsection{Net2net}\label{app:net2net}

\textcolor{black}{We use Net2Net~\citep{chen2016net2net} for expanding a base model which results in a functionally equivalent model. To increase width, its Net2WiderNet operation expands a layer by adding new neurons or channels, initializing their weights as duplicates of existing ones and adjusting the downstream weights accordingly to preserve the output. To add new layers, its Net2DeeperNet operation adds new layers initialized as identity functions, ensuring they do not change the output of the network at initialization. These function-preserving transformations allow the larger model to produce the same outputs as the smaller one before training, which eliminates the need for retraining from scratch and leads to faster and more stable convergence. Unlike zero initialization, which starts the expanded model from a relatively uninformative state, Net2Net provides a well-informed starting point that avoids relearning and results in significantly faster convergence and more stable training.}

\subsection{Baselines}\label{app:baselines}

We compare our manifold metric with the following NAS baselines used in~\citep{white2021powerful}. Our implementation of these baselines in Jax is based on the code made available by~\citep{mellor2021neural}.% that used them for NAS. We describe these baselines as follows:
\begin{itemize}
    \item GradNorm~\citep{abdelfattah2021zero}: It is the sum of the Euclidean norm of the gradients of one minibatch of training data. This measure is an indication of the property of loss landscape.
    \item Jacov~\citep{mellor2021neural}: It is the covariance of the Jacobian of the model's prediction with respect to the input. It involves computing the highest eigenvalue of the covariance matrix. 
    \item SNIP~\citep{lee2018snip}: Single-shot network pruning (SNIP) is a network pruning method that involves masking the parameters and running a backward pass.
    \item Grasp~\citep{wang2020picking}: Gradient signal preservation is a network pruning method that involves computing a Hessian vector product. 
    \item SynFlow~\citep{tanaka2020pruning}: Synaptic Flow is another network pruning method that involves a backward pass and a vector-vector product computation. {Its main advantage over other zero-cost proxies is the fact that it generally avoids layer collapse when performing parameter pruning by taking a product of the parameters in the network \cite{white2021powerful}.}
    \item SoTL-E~\citep{ru2021speedy}: It is the sum of the training losses in the most recent epoch. Unlike other baselines which are zero-cost proxies, it requires training the model and is therefore significantly more expensive. 
\end{itemize}

% All the above baselines are zero-cost proxies except SoTL-E. We compare our manifold metric with the zero-cost proxies based on computations involved in \autoref{tab:baselines}. We also report the time taken by each metric on a checkpoint just after expansion (with the highest expansion factor) from different image classification setups. As the manifold metric does not require running a backward pass and primarily requires running multiple forward passes, it is computationally more efficient than the baselines. We observe the same in comparing the time taken when the model complexity increases from CNN to ResNet18. More importantly, the manifold metric is faster than the second-best baseline SynFlow. 

\autoref{tab:baselines} compares our manifold metric against the other above-mentioned zero-cost proxies. The metrics efficiency is measured as the time (in seconds) to compute them on a checkpoint immediately after expansion with the highest factor for different image classification setups. Since the manifold metric does not require a backward pass but rather performs multiple forward passes, it is computationally more efficient than the baselines. The efficiency holds when the model complexity increases from CNN to ResNet18. Notably, the manifold metric is faster than the second-best baseline, SynFlow.

\begin{table*}[h!]\caption{Comparison between our manifold metric with zero-cost baselines based on computation involved and the actual time taken (in seconds) on a given checkpoint using a single batch of data. \\
$*$ indicates that gradients are computed on outputs instead of loss.}\label{tab:baselines}
\centering
\resizebox{\textwidth}{!}{
\begin{tabular}{@{}cccccc@{}}
\toprule
\textbf{Method} & \textbf{Number of} & \textbf{Highest}  & \multicolumn{3}{c}{\textbf{Time taken (seconds)}} \\ 
 & \textbf{forward passes} & \textbf{derivative order}  &  &  & \\ 
 &  &   & {CNN (CIFAR10)} & {CNN (CIFAR100)} & {ResNet18} \\ 
\midrule
{GradNorm} & $1$ & $1$ & $\textbf{0.4}_{\pm 0.01}$ & $\textbf{0.5}_{\pm 0.01}$ & $4.4_{\pm 1.1}$ \\
{Jacov} & $1$ & $1^*$ & $\textbf{0.4}_{\pm 0.01}$ & $0.6_{\pm 0.01}$ & $2.8_{\pm 0.7}$ \\
{SNIP} & $1$ & $1$ & $\textbf{0.4}_{\pm 0.02}$ & $0.5_{\pm 0.01}$ & $3.8_{\pm 1.0}$ \\
{Grasp} & $1$ & $2$ & $0.5_{\pm 0.01}$ & $0.7_{\pm 0.02}$ & $6.4_{\pm 1.5}$ \\
{SynFlow} & $1$ & $1^*$ & $0.6_{\pm 0.11}$ & $0.7_{\pm 0.13}$ & $4.4_{\pm 1.2}$ \\
\midrule
{Manifold (ours)} & $2n+1~(=101)$ & $0$ & $\textbf{0.4}_{\pm 0.01}$ & $0.6_{\pm 0.01}$ & $\textbf{2.1}_{\pm 0.5}$ \\
\bottomrule
\end{tabular}}
\end{table*}

\subsection{Correlation Metrics}\label{app:corr_met}

\begin{figure*}[htb!]
    \centering
    \includegraphics[width=\textwidth]{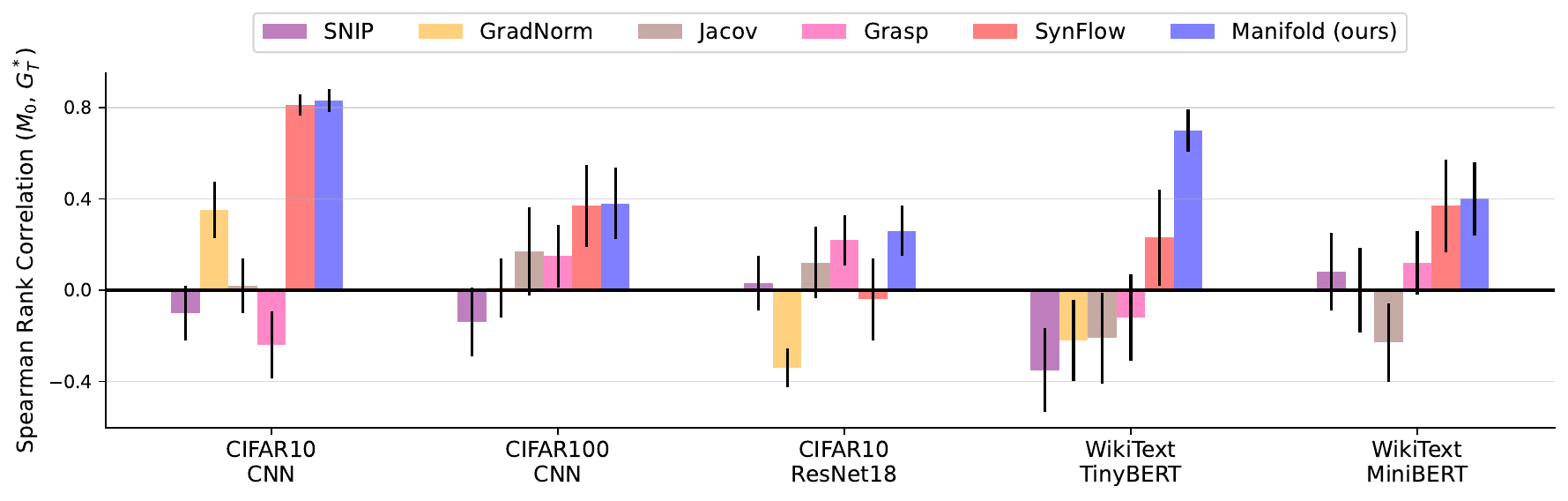}\\
    \includegraphics[width=\textwidth]{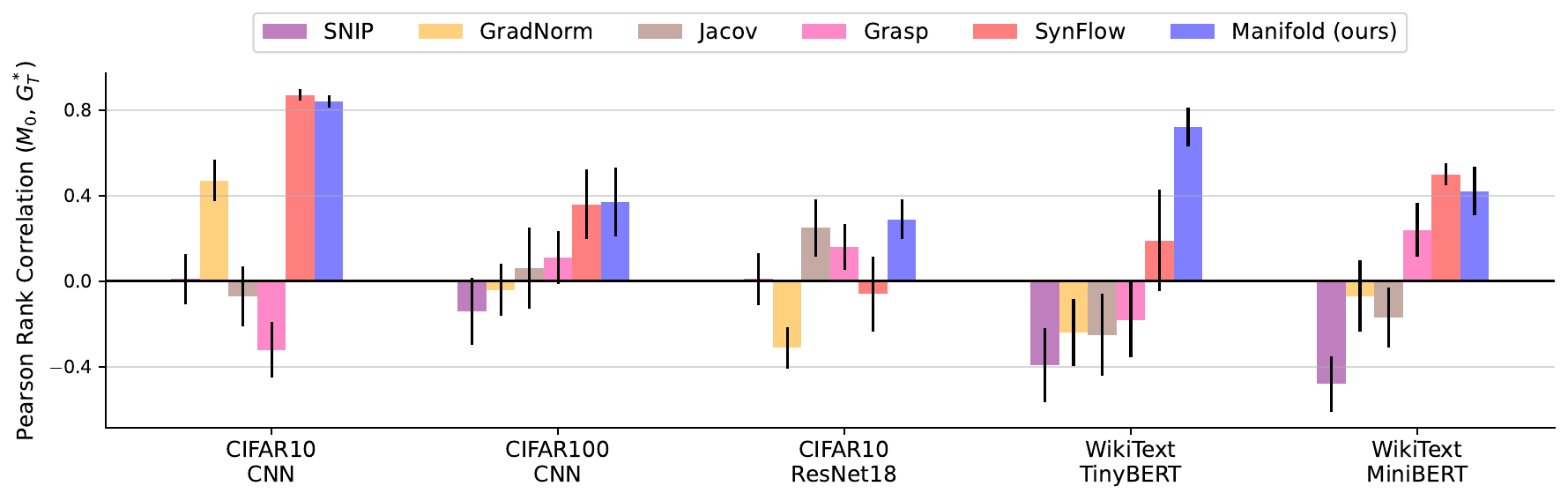}
    \includegraphics[width=\textwidth]{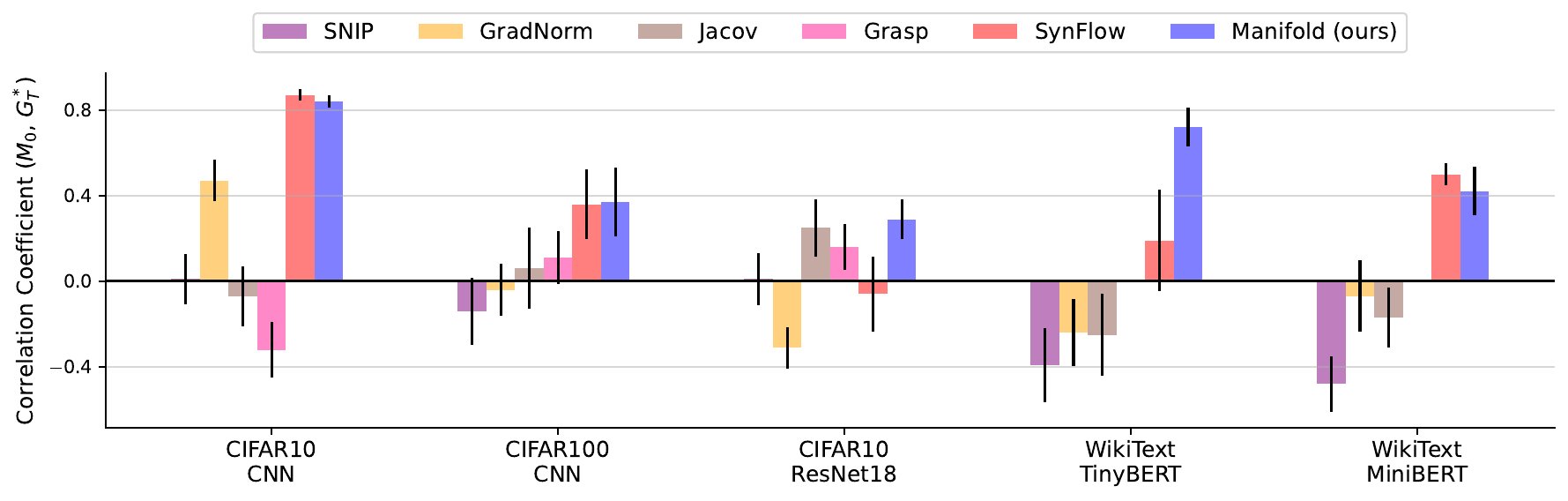}
    \caption{Comparison between the manifold metric $M_0$ and the existing zero-cost proxies in terms of Spearman rank correlation (first) and Pearson rank correlation (second)  and correlation coefficient (third) with $G_T$ (final performance gain). The horizontal axis notes the dataset and the base model. Correlation by manifold metric remains consistently positive, whereas baselines exhibit both positive and negative correlations similar to \autoref{fig:all_tau}.}% except BERT with self-attention (SA) layer expansion.}
    \label{fig:all_rank}
\end{figure*}

In this section, we provide a comparison between the manifold metric and the zero-cost baselines in terms of other correlation metrics. We plot the results in \autoref{fig:all_rank}. In each case, we observe the same trend as \autoref{fig:all_tau} where, unlike the baselines, the correlation between the manifold metric and the performance gain remains positive across different models and datasets, and other baselines.

\subsection{Sensitivity Analysis}\label{app:sens}

% \subsection{Sensitivity Analysis}\label{sec:sens}

This section presents sensitivity analysis results obtained by performing a grid search over the sets $q$ and $n$. The goal of such an analysis is to show the robustness of our manifold metric $M_0$ across different optimization settings and models by calculating the Kendall Tau correlation with the final performance gain $G^*_T$ for various combinations of $(q, n)$. %\autoref{fig:sens_mnist},

\begin{figure*}[htb!]
    \centering
    \resizebox{\linewidth}{!}{
        \includegraphics[width=0.3\textwidth]{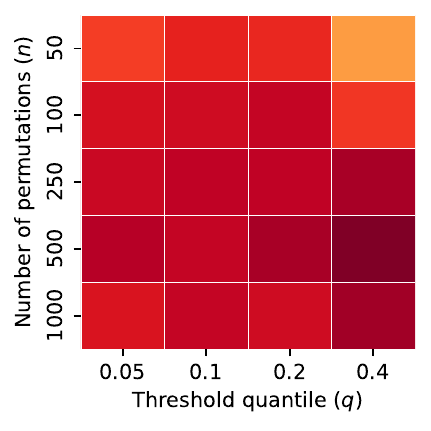}
        \includegraphics[width=0.3\textwidth]{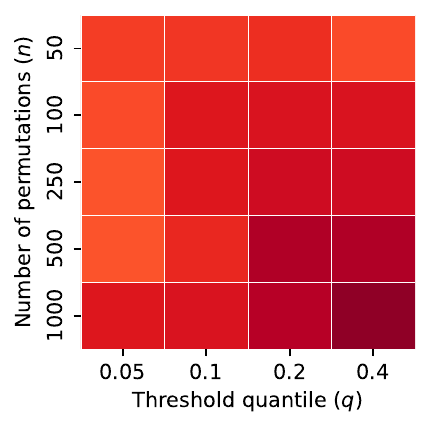}
        \includegraphics[width=0.3\textwidth]{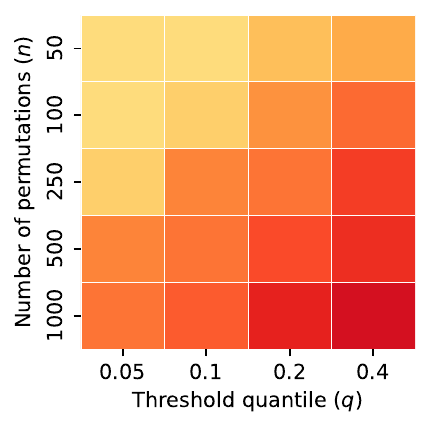}
        \hfill
        \includegraphics[width=0.057\textwidth]{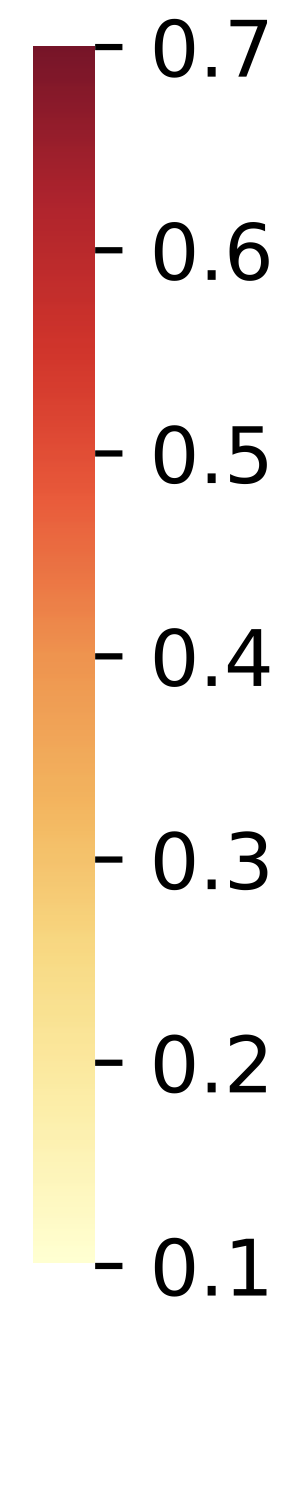}
    }
    \caption{Sensitivity analysis on CNN on CIFAR10 with Adam (first), AdamW (second) and TinyBERT on WikiText-103 with AdamW (third), in terms of Kendall Tau correlation observed between the manifold metric $M_0$ and the overall performance gain $G^*_T$. We observe that correlation generally increases as we grow the number of permutations $n$ and threshold quantile $q$.}
    \label{fig:sens_image_new}
\end{figure*}

\autoref{fig:sens_image_new} illustrates the average correlation obtained over $10$ different seeds for the following settings: expanding a CNN model pre-trained on CIFAR10 using Adam (first), AdamW (second), and expanding a TinyBERT model pre-trained on Wikitext-103 using AdamW (third). For all combinations of $(q, n)$ in all three cases, the manifold metric results in a positive correlation. More importantly, higher values of $n$ result in higher correlation, which suggests that more accurate estimates of the manifold size can indeed lead to better correlation with the performance gain. Additionally, for both CNN and TinyBERT models pre-trained using AdamW, a higher value of $q$ tends to result in better correlation. A similar pattern occurs in CNN with Adam, despite less pronounced trends, where a smaller $q=0.05$ also results in a high correlation.

\begin{figure}[!htb]
    \centering
    \includegraphics[width=0.32\textwidth]{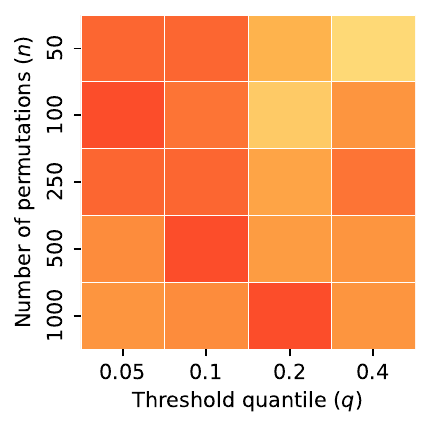}
    \includegraphics[width=0.32\textwidth]{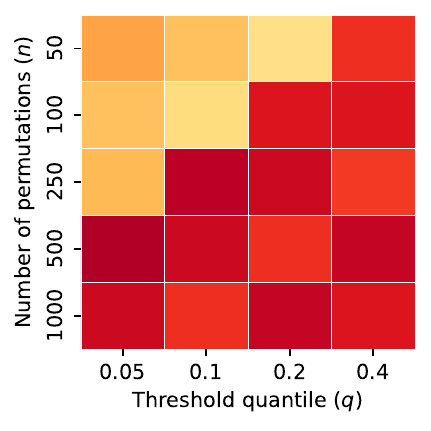}
    \includegraphics[width=0.061\textwidth]{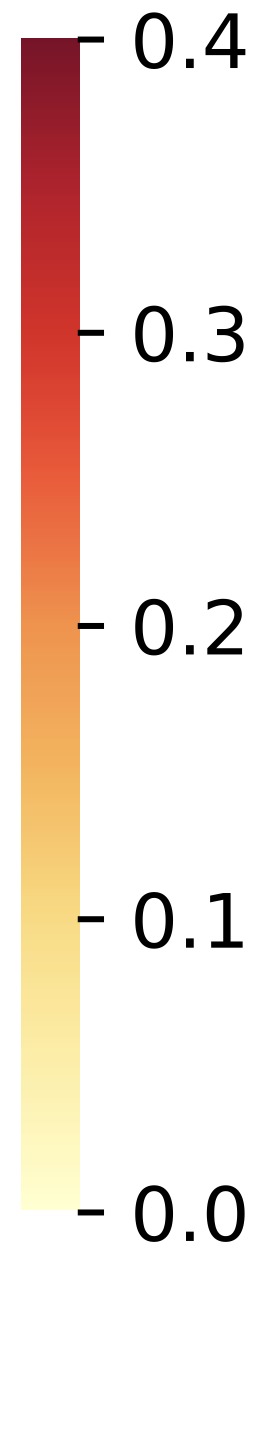}
    \caption{Sensitivity analysis on CNN on CIFAR100 (first) and MiniBERT on WikiText-103 (second), in terms of Kendall Tau correlation observed between the manifold metric $M_0$ and the overall performance gain $G^*_T$. We observe that the correlation is low but positive for CIFAR100. On the other hand, it increases as we grow the number of permutations $n$ and threshold quantile $q$ in MiniBERT.}
    \label{fig:sens_app_image}
\end{figure}

\autoref{fig:sens_app_image} further illustrates the average correlation obtained from ten seeds for CIFAR100 and MiniBERT experiments. We observe that there is not any clear trend in the case of CIFAR100 due to low overall correlation, unlike CIFAR10 in \autoref{fig:sens_image_new}. On the other hand, the MiniBERT correlation remains high for a larger number of permutations $n$ and threshold quantile $q$ in the case of MiniBERT.

Overall, these findings suggest that the change in manifold metric serves as a reliable estimator of potential performance gain when expanding a pre-trained model. Furthermore, a high optimal $q$ indicates that this estimation does not rely explicitly on the linearity of connections between minima. Rather, as long as the minima are connected through low-loss paths, the change in manifold size can predict the impact of expansion on overall performance.

\subsection{Training CNN with Adam}\label{app:adam}
Earlier in \autoref{sec:image}, the results were obtained when the expanded CNN models were trained using the AdamW optimizer. In this section, we run similar experiments but train the CNN models using the Adam optimizer to investigate how weight decay affects the manifold of the loss landscape during the training process.

\begin{figure*}[t]
    \centering
    \resizebox{\linewidth}{!}{
        \includegraphics[width=0.64\linewidth]{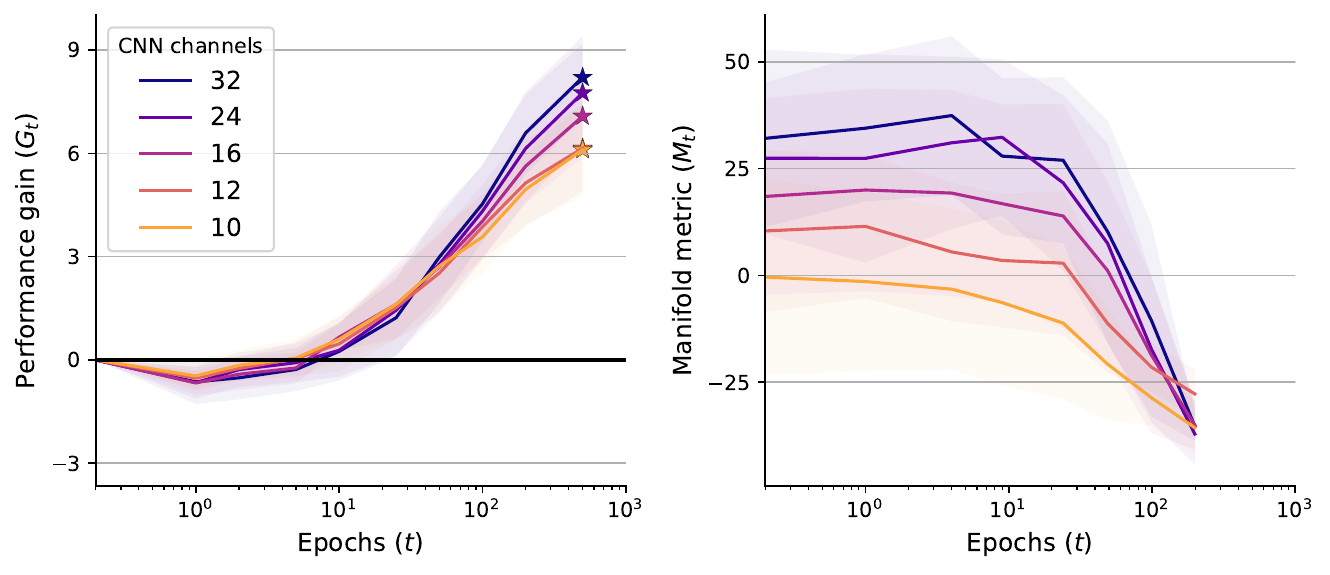} 
    	\includegraphics[width=0.35\linewidth]{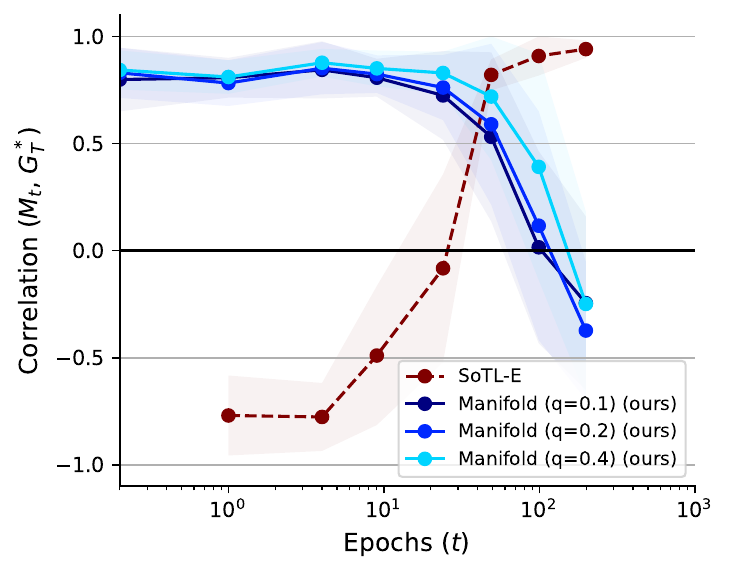}
    }
    \resizebox{\linewidth}{!}{
        \includegraphics[width=0.64\linewidth]{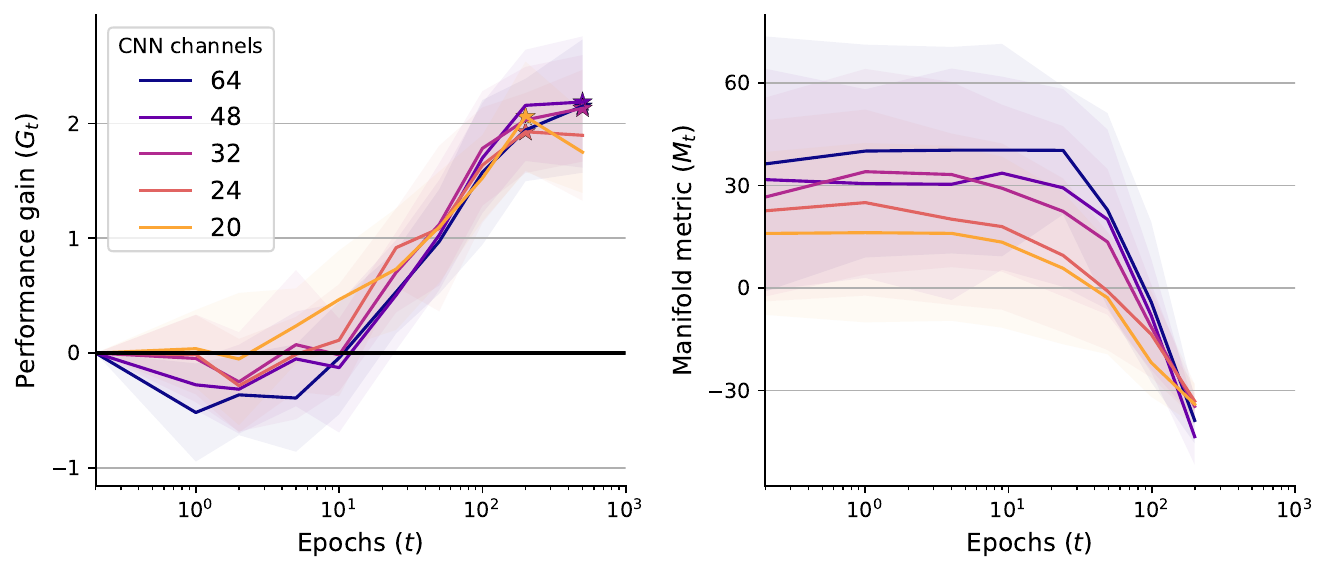}
    	\includegraphics[width=0.35\linewidth]{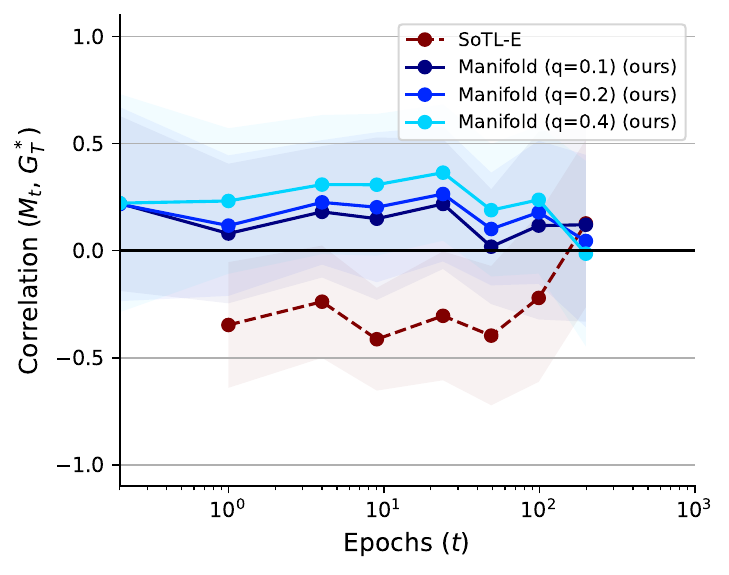}
    }
    \caption{Performance gain $G_t$ (first column; $\star$ denotes the best performance gain $G^*_t$), manifold metric $M_t$ (second column) and correlation between $M_t$ and the highest gain $G^*_T$ (third column) on expanding and training CNN on CIFAR10 (top) or CIFAR100 (bottom) for $T=500$ epochs when a model is pre-trained using Adam optimizer. Similar to \autoref{fig:image_manifold}, $M_t$ remains constant and proportional to $G^*_T$ from the beginning and drops during the later stages of training.} 
    \label{fig:image_manifold_app}
\end{figure*}

\begin{figure*}[htb!]
    \centering
    
        \includegraphics[width=0.32\linewidth]{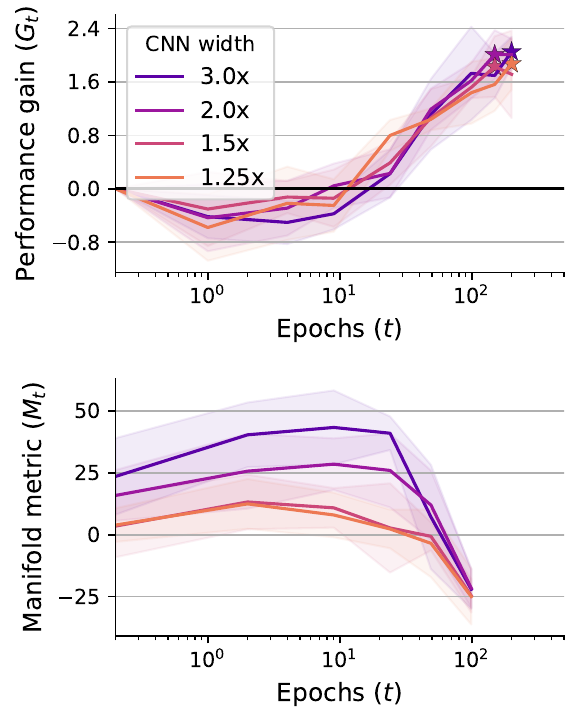} 
        \includegraphics[width=0.32\linewidth]{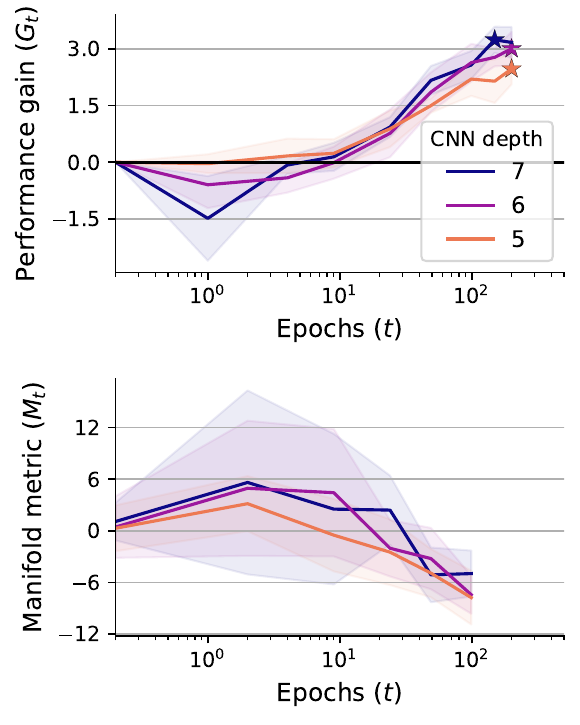}
    \caption{Performance gain $G_t$ (first row; $\star$ denotes the best performance gain $G^*_t$) and manifold metric $M_t$ (second row) for CNN width expansion for all layers (first column) and CNN depth expansion on CIFAR100 (second column) using Adam optimizer. The manifold metric $M_t$ positively correlates with the level of expansion initially. However, we also observe a quicker drop for $M_t$ in case of depth expansion.} 
    \label{fig:image_manifold_ablation_app}
\end{figure*}

The corresponding results shown in \autoref{fig:image_manifold_app} and \autoref{fig:image_manifold_ablation_app} have a similar trend as the AdamW experiments. However, we also observe a quicker drop for $M_t$ in case of large expansion factors during the later training phase. Thus, we conclude that the manifold metric is robust toward the use of weight decay in capturing geometric properties of the loss landscape, especially in the initial phase which is more desirable to evaluate model expansion.

\begin{figure}
    \centering
    \includegraphics[width=0.6\linewidth]{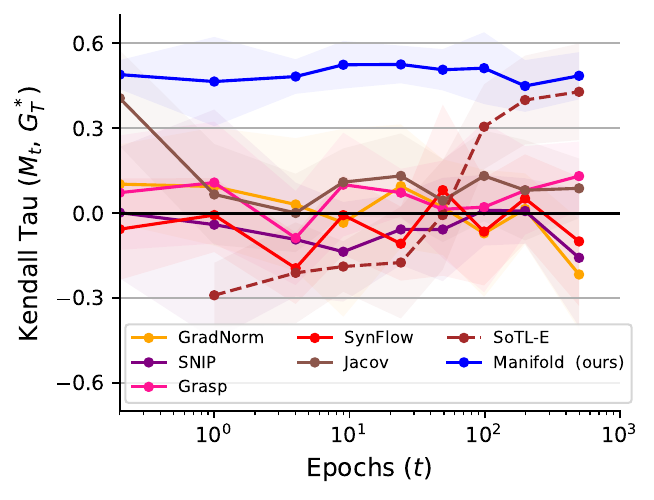} 
    \caption{Comparing rank correlation between different metrics and the highest gain $G^*_T$ observed after expanding (maintaining similar number of parameters) and training CNN models on CIFAR10. Overall, our manifold metric correlates better with $G^*_T$ throughout training, making it more reliable for comparing expanded models. All other baselines have poor correlation, except Jacov, which has the second-best correlation after expansion.}
    \label{fig:same_size_adam}
\end{figure}

We also conduct a similar ablation study, as described in \autoref{sec:ablation}, by expanding a CNN model in such a way that the number of parameters remains similar. We then train the candidate models using the Adam optimizer. \autoref{fig:same_size_adam} shows similar trends, observed in previously in \autoref{fig:same_size}, where our manifold metric $M_t$ significantly outperforms other baselines in terms of rank correlation throughout training.

% \begin{wrapfigure}{r}{0.4\textwidth}
%  % \vspace{-55px}
% \begin{center}
% \end{center}
% \vspace{-35px}
% \end{wrapfigure}

\subsection{Online learning}\label{app:online}

\textcolor{black}{In this experiment, we evaluate our manifold metric on online learning setups. While we focused on the same task setup in our main experiments, the goal of this experiment is to verify whether manifold metric can still perform better than other zero-cost proxies under more complex settings that involve realistic distribution shift. We use a sequence of 5 tasks. The first task is trained using the default CNN model used for CIFAR10 (see \autoref{tab:models_image}). After training on a task, we evaluate candidate expansions and use a zero-cost proxy to rank and select the best model for the next task. We expand the model with the highest-ranked candidate and train it with early stopping. The goal is to maximize final task generalization performance. We evaluate on two types of benchmarks:}
\begin{itemize}
    \item \textcolor{black}{Data Curricula: Each task is an increasingly larger subset of a given dataset (e.g., $20\%, 40\%, …, 100\%)$. }
    \item \textcolor{black}{5-Dataset: Sequence of 5 different datasets—MNIST, FashionMNIST, SVHN, CIFAR-10, and STL-10.}
\end{itemize}
\textcolor{black}{In all cases, we used AdamW(lr=0.001). We report the performance averaged across 3 seeds in \autoref{tab:cl}. }

\begin{table}[htb!]
\centering
\caption{\textcolor{black}{Comparison Final Task Accuracy (in \%) obtained using model selection in online learning.}}\label{tab:cl}
\begin{tabular}{lccc}
\toprule
\textbf{Method} & \textbf{Data: CIFAR-10} & \textbf{Data: CIFAR-100} & \textbf{5-Dataset} \\
\midrule
No Expansion     & 72.5 & 45.9 & 36.0 \\
\midrule
Grasp            & 74.3 & 46.4 & 37.6 \\
SNIP             & 73.1 & \textit{47.1} & 41.2 \\
Jacov            & 74.5 & 46.5 & \textit{41.6} \\
SynFlow          & \textit{76.3} & 46.4 & 37.6 \\
\textbf{Manifold (Ours)} & \textbf{76.8} & \textbf{47.9} & \textbf{43.4} \\
\bottomrule
\end{tabular}
\end{table}
\textcolor{black}{Overall, we observe that the Manifold metric outperforms other baselines on all three benchmarks, demonstrating both robustness and generalization. Interestingly, the gap between the Manifold metric and the second-best baseline is larger in the 5-dataset setting.
}

\subsection{Additional Experiments for Transformer}\label{app:transformer}

We provide the results of ablation experiments for analyzing how mode connectivity changes in a Transformer.

\begin{figure}[!htb]
    \centering
    \includegraphics[width=0.48\linewidth]{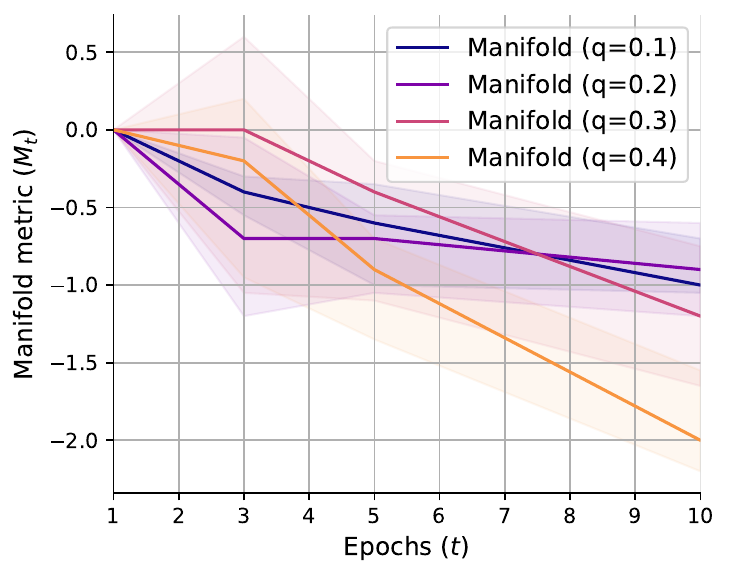}
    \hfill
    \includegraphics[width=0.48\textwidth]{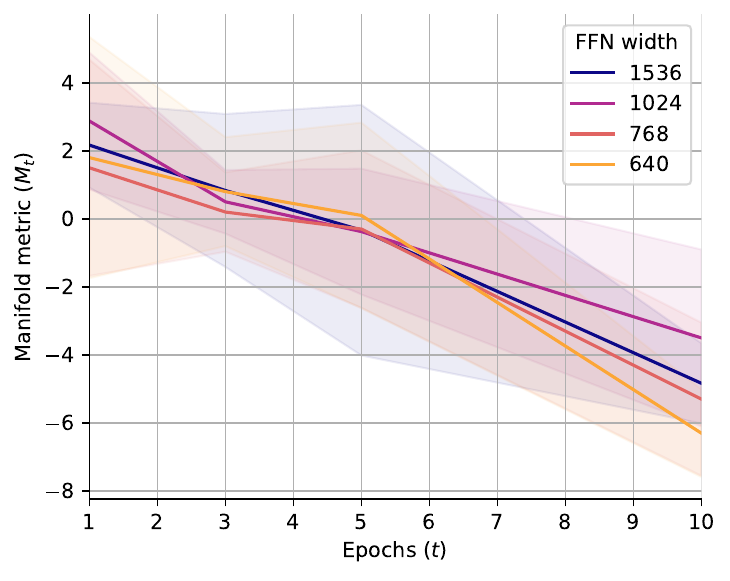}
    \caption{(i) Comparing manifold metric while training TinyBERT on WikiText-103 for $T=10$ epochs starting from random initialization. Similar to \autoref{fig:BERT_manifold}, the manifold metric drops as the model is trained further. (ii) Manifold metric $M_t$ computed by permuting self-attention weights during training TinyBERT model on WikiText dataset. The manifold metric decreases as the model is trained similar to \autoref{fig:BERT_manifold}.}
    \label{fig:BERT_random}
\end{figure}

We train a Transformer from a random initialization for $10$ epochs. The manifold size at initialization should be vast due to parameters lying on a high-loss surface. Therefore, to compute $M_t$, we use the mode connectivity at epoch $1$ as a reference point and monitor how it changes during training. We compare the metric $M_t$ across different $q$ and plot them in \autoref{fig:BERT_random} (first). We observe that in all cases, the mode connectivity reduces. Moreover, the drop grows with $q$, suggesting that even the non-linear mode connectivity between parameters drops as the model is trained. 

In another experiment, we permute self-attention layer weights instead of FFN weights of the first encoder layer in TinyBERT to obtain different minima. \autoref{fig:BERT_random} (second) shows the manifold metric obtained during training different expanded models. We observe that the manifold metric decreases throughout training similar to \autoref{fig:BERT_manifold} (middle). This ablation experiment also suggests that the manifold metric does not depend on the type of permutation applied to this model to obtain different minima.

\end{document}